\documentclass[pdftex,11pt,openright,headsepline]{book}

\usepackage{paralist}		% List environment
\usepackage{color}		% For colored text
\usepackage{times}
\usepackage{amsfonts}		% Additional math fonts
\usepackage{amsmath}		% Math symbols
\usepackage{latexsym}
\usepackage{graphicx}		% For including images
\usepackage{mydefs}		% Some of our own definitions
\usepackage{fancyhdr}		% Produce the nice header
\usepackage{fullpage} % Use the full page
\usepackage{hyperref}

\usepackage[headsep=0.3cm,headheight=1cm]{geometry}
\begin{document}
\pagestyle{empty} % even no page number

\fancypagestyle{plain}{
  \renewcommand{\headrulewidth}{0.0pt}
  \fancyfoot{}
  \fancyhead{}
}

% Title page, modify accordingly 
%% ----------------------------------------------------------------------------
% BIWI SA/MA thesis template
%
% Created 09/29/2006 by Andreas Ess
% Extended 13/02/2009 by Jan Lesniak - jlesniak@vision.ee.ethz.ch
%% ----------------------------------------------------------------------------

\begin{titlepage}

\thispagestyle{empty}

\fancypagestyle{empty}{
\lhead{\includegraphics[height=1.5cm]{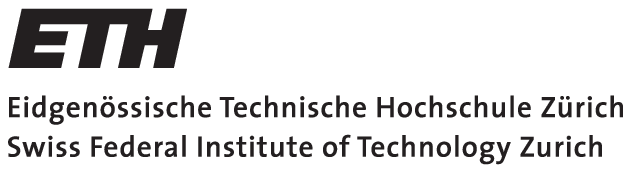}}
\renewcommand{\headrulewidth}{0.0pt}
\rhead{\vspace*{-0.2cm}\includegraphics[height=1.4cm]{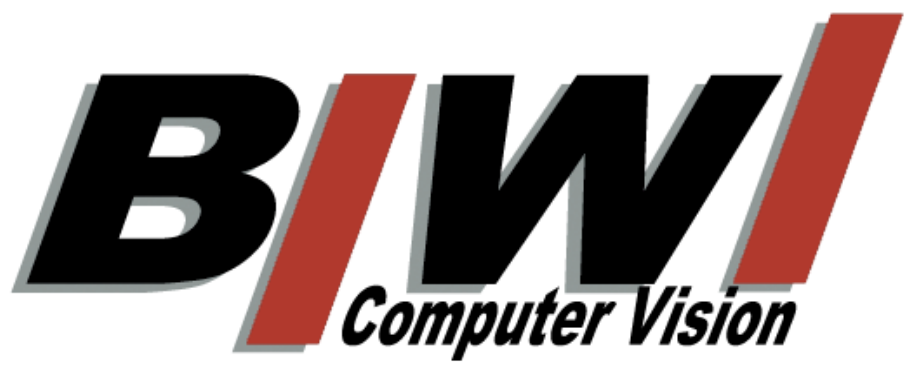}}
\fancyfoot{}
}

\vspace*{2cm}
\begin{center}
\Huge{\textbf{Real-time 3D Pose Estimation with a Monocular Camera Using Deep Learning and Object Priors}\\}
\LARGE{\textbf{On an Autonomous Racecar}\\[1cm]}

\large{Semester Thesis\\[0.8cm]}
\LARGE{Ankit Dhall\\}
\normalsize{Department of Mechanical and Process Engineering}
\end{center}

\begin{center}

% \begin{center}
% \begin{tabular}{ll}
% \multirow{2}{*}{\includegraphics[height=1cm]{images/biwi_logo}} & Computer Vision Laboratory\\ 
% & ETH Zurich
% \end{tabular}
%  \end{center}

\end{center}

\vfill
\begin{center}
\begin{tabular}{ll}
\Large{\textbf Advisor:} & \Large{Dr. Dengxin Dai}\\
\Large{\textbf Supervisor:} & \Large{Prof.~Dr.~Luc van Gool}\\
% 			    & \small{Computer Vision Laboratory}\\
% 			    & \small{Department of Information Technology and Electrical Engineering}\\
\end{tabular}
\end{center}

\begin{center}
\today\\
\end{center}

\end{titlepage}

\cleardoublepage

%% ----------------------------------------------------------------------------
% BIWI SA/MA thesis template
%
% Created 09/29/2006 by Andreas Ess
% Extended 13/02/2009 by Jan Lesniak - jlesniak@vision.ee.ethz.ch
%% ----------------------------------------------------------------------------

\newpage
\vspace{3cm}

\chapter*{Abstract}
\noindent In computer vision, triangulation via arranging two cameras in a stereo setup has become the norm in order to estimate the 3D pose of a particular object of interest and is used in most autonomous robots to help perceive the environment.

In experiments limited to laboratory environments, classical computer vision techniques such as stereo correspondence search and triangulation work decently well. Moreover, one could utilize off-the-shelf equipment such as the Kinect sensor or Intel RealSense. The drawback being that such sensors have limitations when taken outdoor and have only a limited range (up to a few meters). This makes it infeasible to use such setups for long-range pose estimation.

Range and time-of-flight sensors can be used to extract 3D information using raw data provided by such sensors from point clouds. But again, detecting particular objects in such point clouds is non-trivial. Having to do this for multiple objects of interest only compounds the task. Although, time-of-flight sensor manufacturers are trying to cut down costs and make such with competitive prices but are still a long way from manufacturing accurate sensors available at a competitive price such as cameras (which are orders of magnitude cheaper and provide most information per cent). 

A monocular camera, as compared to a stereo or a multi-camera setup, on the other hand has lesser data involved by a factor of number of cameras in the multi-camera setup. Another advantage of using a monocular camera is that one can bypass problems that arise from a stereo setup such as synchronization and stereo calibration. On the other hand, however, estimating 3D pose of an object using a single measurement, i.e. a single image from a monocular camera is an ill-posed problem. This is primarily due to ambiguity in the scale of the scene arising from limited information of the surroundings.

This ill-posed problem of extracting pose information can be solved if a priori information about the 3D object in the scene is available. The 3D priors about an object, in addition to 2D information obtained from an image can be together leveraged to extract 3D pose of this object captured in any arbitrary image of the scene.

On a real-time system, such as an autonomous race-car, it becomes even more crucial to detect and estimate multiple object positions extremely efficiently, with as little latency and data overhead (in terms of transport and processing) as possible. To this end, we propose a low-latency real-time pipeline to detect and estimate 3D position of multiple objects of interest using just a single measurement, i.e. a single image without the need for any special external markers. This allows the perception pipeline to be extremely light-weight (requiring just a single camera) and perform pose estimation for multiple objects of interest in the wild using a single image.

We propose a novel ``keypoint regression" scheme that exploits prior information about the object's shape and size to regress and find specific feature points on the image. Further, in addition to the ``keypoint regression" which provides 2D information, a priori 3D information about the object is used to match 2D-3D correspondences. We propose a complete pipeline that allows object detection and simultaneously estimate pose of these multiple object using just a single image by exploiting object priors. As a design choice, we employ a complementary part-based approach that uses a combination of data-driven deep learning, exploiting data wherever required, and at the same time employs results from classical computer vision at places where machine learning and deep learning would be an overkill. This makes our pipeline more intuitive, appealing and easy-to-debug as compared to a hypothetical end-to-end deep learning pipeline.

To demonstrate the quality of our pose estimates and the system as a whole, we deploy it as part of the perception pipeline on ``gotthard driverless". ``gotthard driverless" competed in an international, highly competitive engineering competition spanning over 9 months called Formula Student Driverless. The light-weight aspect of a monocular camera makes it efficient and minimizes both computation resources usage and data overhead. We show how using the proposed pipeline one can dramatically increase the perception range without losing on accuracy of position estimates, translating to direct on-track performance of the autonomous race car.

With a combination of a clever choice of optics and the ``keypoint regression", the stand-alone monocular camera can perceive cone position and color up to 15m. The pipeline runs in real-time on onboard computers and serves as input to prepare a map of an unknown track by detecting and classifying colored cones and their 3D position on a per-frame basis with very low latency. The cones help the car plan its trajectory through the track. As part of AMZ Driverless 2018, ``gotthard driverless" participated in Formula Student Italy (FSI) and Formula Student Germany (FSG) 2018, the biggest engineering competition in the world with 100+ teams from all over the world. We managed to achieve the record highest score ever in a Formula Student event, 1000/1000 points (at FSI) and record highest score by any team at FSG, 959/1000 points. In both the competitions it was declared as the overall Champion, cruising at a top speed of 54kmph autonomously.

% The abstract gives a concise overview of the work you have done. The reader shall be able to decide whether the work which has been done is interesting for him by reading the abstract. Provide a brief account on the following questions:

% \begin{itemize}
%  \item What is the problem you worked on? (Introduction)
%  \item How did you tackle the problem? (Materials and Methods)
%  \item What were your results and findings? (Results)
%  \item Why are your findings significant? (Conclusion)
% \end{itemize}

% \noindent The abstract should approximately cover half of a page, and does generally not contain citations.

% Input here any acknowledgements
%% ----------------------------------------------------------------------------
% BIWI SA/MA thesis template
%
% Created 09/29/2006 by Andreas Ess
% Extended 13/02/2009 by Jan Lesniak - jlesniak@vision.ee.ethz.ch
%% ----------------------------------------------------------------------------

\newpage

\chapter*{Acknowledgements}

I would like to thank AMZ Driverless for this amazing opportunity to work on a life-sized, real-time and very demanding yet extremely interesting project. Not only did it help me learn new skills and improve my technical abilities but also helped me sharpen my soft skills and grow on a personal level.

To make the computer vision part of perception a success on ``gotthard driverless" it would not have been possible if it was not for the effort and long hours of discussion about the concept with Mehak and Ramya. Making sure we had a reliable platform to run the algorithms on and ensuring fully-functioning cameras was not an uphill task and with their help we managed to overcome it. Mehak's efforts on the stereo pipeline gave us the ability to provide position estimates with pin-point accuracy (for nearby cones), making the whole package very reliable. I would also like to thank her for helping me review my presentation and giving constructive suggestions.

I would also like to thank Miguel for guiding us and Juraj for helping translate this into a running pipeline on-board. Nicolo's assistance, for all things small and big, made it quite easy to run our pipeline on the computing system of ``gotthard driverless".

Thanks to Lisa for helping us arrange sponsoring from Basler for the cameras and optics and Jochen for bringing life into the whole camera housing package with an sleek and futuristic design. I would like to thank my team members for their patience and belief that we could build ``gotthard driverless" and make it drive autonomously despite all the hiccups we faced throughout the season.

Thanks to Susanna and Giuseppe for keeping my spirits high and being there whenever I needed to boost my morale. Also, this would have not been the same if it wasn't for their awesome late-night cooking.

Finally, I would like to thank the ``Computer Vision Lab" and especially, Dengxin and Prof. Luc van Gool for providing me with the opportunity to pursue my ideas with ``gotthard driverless" as a project with them. Dengxin's feedback was quite crucial to make this a successful endeavour. Lastly, thanks to my parents and my brother for always being understanding and supportive throughout this long and arduous project.

\cleardoublepage
\newpage

% % Chapter-pages etc. use the ``plain'' pagestyle - since we don't want to have a heading at all at chapter-pages, redefine plain accordingly. Don't forget the page number. 
\fancypagestyle{plain}{
  \renewcommand{\headrulewidth}{0.0pt}
  \fancyfoot{}
  \fancyfoot[RO, LE]{\thepage}
  \fancyhead{}
}

\pagestyle{fancy}
\pagenumbering{Roman}

% Insert table of contents
\tableofcontents

% Insert list of figures
\listoffigures
\cleardoublepage

% Insert list of tables
\listoftables
\cleardoublepage

\newpage

\pagenumbering{arabic}

%% ----------------------------------------------------------------------------
% Actual text comes here - defer it to other files and use \input{bla.tex}, ..
%% ----------------------------------------------------------------------------
%% ----------------------------------------------------------------------------
% BIWI SA/MA thesis template
%
% Created 09/29/2006 by Andreas Ess
% Extended 13/02/2009 by Jan Lesniak - jlesniak@vision.ee.ethz.ch
%% ----------------------------------------------------------------------------

\chapter{Introduction}

\section{Formula Student Driverless - Why we built ``gotthard driverless"?}

Before we delve into details about the project it is important to get an overview of the bigger picture and why this project was undertaken in the first place. The idea of building an autonomous race car is to compete in a 9-month long engineering competition where students teams from across the world strive to build the fastest, smartest and the most reliable autonomous race cars and compete head-to-head in week long events.

\begin{figure}[h]
    \centering
    \includegraphics[width=1.0\textwidth]{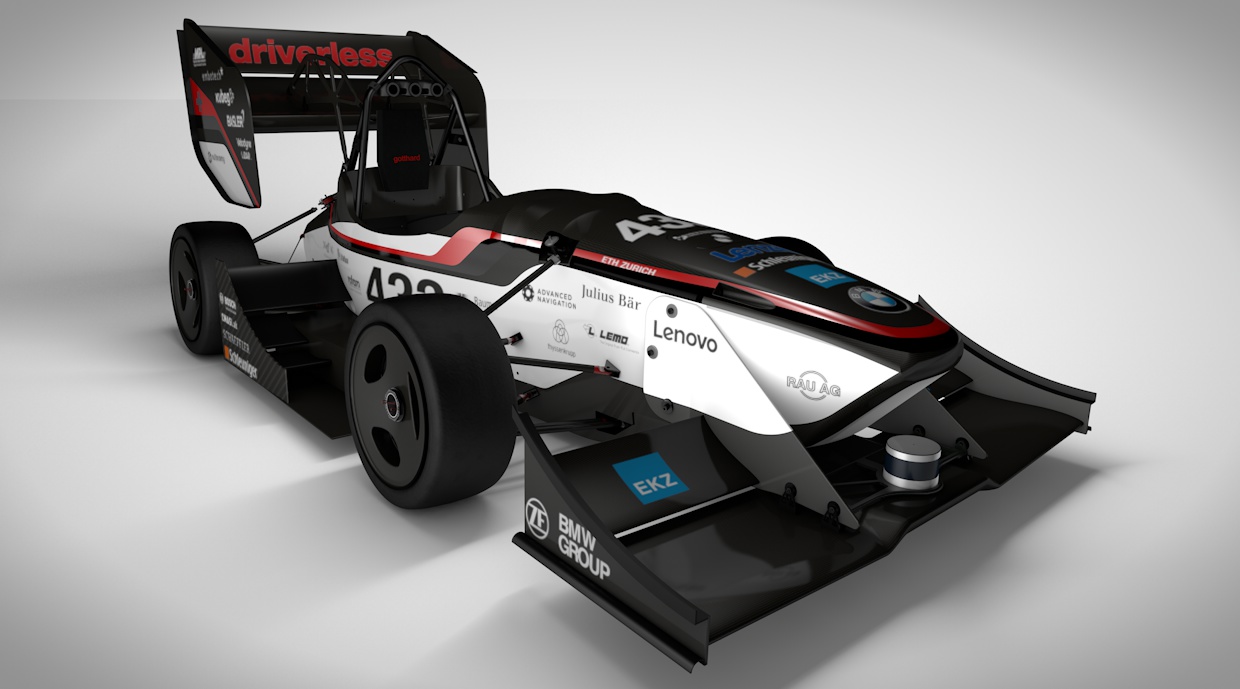}
    \caption{``gotthard driverless", 4-wheel driven, autonomous electric race car with self-developed lightweight and efficient motors. The LiDAR can be seen under the nose and the cameras are mounted above the driver's seat on the roll hoop.}
    \label{fig:gotthard_rendering}
\end{figure}

The events are categorized as statics and dynamics. The statics comprise of ``Cost Presentation", ``Business Plan" and the most prestigious static event, ``Engineering Design". The dynamic events include ``Acceleration". ``Skid Pad" and ``Trackdrive".

During the ``Cost Presentation" one presents the budget for the car, manufacturing processes their expenses down to every screw and bolt on the car. The aim is to see how one could transition from a prototype to a large-scale production setup. The ``Business Plan" deals with pitching an idea to hypothetical investors about a viable business plan (such as a start-up) distilled from the developments of the last 9 months of working on building the autonomous car. The Engineering Design event is the most important with 325 points up for grabs. Here, the team defends the concept of the car and the design choices that went on to be implemented on the car. It covers both the engineering aspect such as mechanical, electrical and electronics sub-systems on the car and the autonomous system that we build from the ground up.

\begin{figure}[h]
    \centering
    \includegraphics[width=0.5\textwidth]{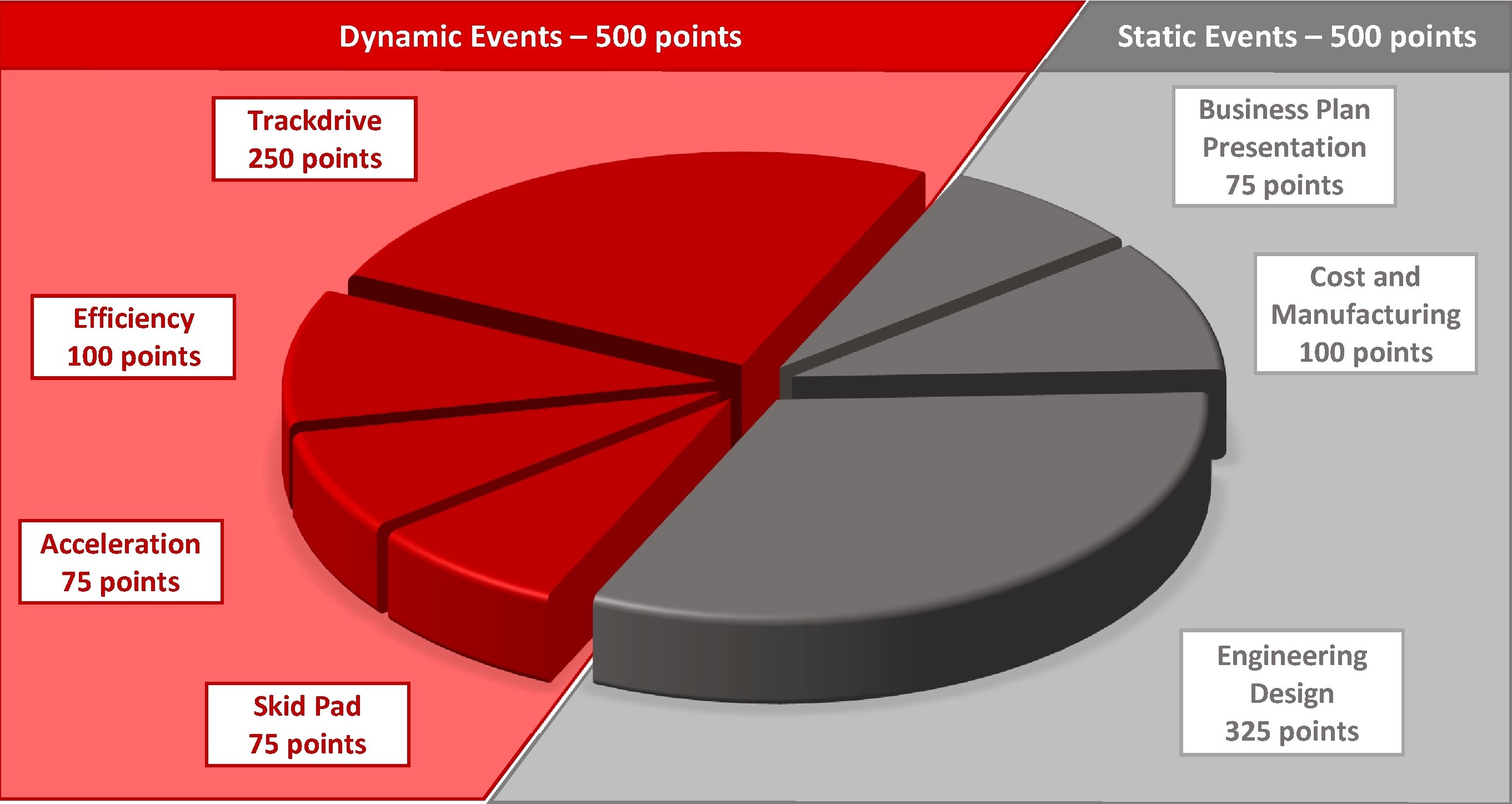}
    \caption{Points distribution during a Formula Student Driverless Event. Usually, the events span across 6-7 days and are categorized as dynamics and statics.}
    \label{fig:points_dist}
\end{figure}

The dynamic events are more about on-track performance. The goal of the ``Acceleration" event is to be the fastest car on a 75 meter straight and brake autonomously within the specified zone. ``Skid Pad" tests the cornering abilities and stability where the car needs to maneuver itself on a circular trajectory (on an ``8" shaped track), two circles on the right followed by two circles on the left. Finally, the most important dynamic event is the ``Trackdrive" where the car is positioned at the start/finish line of an unknown track of about 300-350 meters and needs to complete 10 laps as fast as possible. This is the most crucial dynamic event, worth 350 (=250+100, including efficiency) points and it is where the real pedigree of the perception, mapping and control modules of the autonomous system is put under the pump.

\section{``gotthard driverless'' - 2x Formula Student Driverless Champion}

To build the autonomous system, we use ``gotthard'', an electric car built by AMZ in 2016. ``gotthard'' is a 4 wheel driven electric car with design choices made on the basis of data from previous AMZ cars and self-developed lap-time simulator (LapSim). It features a lightweight device, with a one-piece carbon fibre reinforced plastic (CFRP) monocoque, a full aerodynamics package and custom developed tires. The features of the car enable it to achieve superior lateral, longitudinal and yaw accelerations. The car weighs only 182 kilograms with a top speed of 117 kmph. Until Formula Student Germany 2018, ``gotthard'' drove about 50 kilometers autonomously during testing days.

``gotthard driverless'' perceives its surroundings using raw data from 3 cameras and a LiDAR. To accurately estimate the movement of the car on the track, multiple sensors such as an absolute speed sensor, inertial navigation system and wheel speed sensors are used. Information from these is filtered through to accurately estimate the car's state. A simultaneous mapping and localization (SLAM) module receives low-latency landmark observations from perception modules and fuses them with the car's state to actively create a map of cones. Landmarks on the map are used to extract track limits and find the most plausible path to drive. After completing the first lap and perceiving the  track for the first time, the map is frozen and no longer updated. The module then switches to a localization-only functionality. During the first lap, ``gotthard driverless'' employs a relatively simple pure pursuit controller and does that at slow speeds to ensure mapping accuracy. As soon as the loop is closed, the controller switches to non-linear model predictive controller (NMPCC) which generates smoother and better trajectories by optimizing them over a finite time horizon. The car has a steering actuator that can achieve full left to full right in 0.5s. For braking purposes (in addition to the recuperation on the motors), the car is equipped with an actively braking, emergency brake system (EBS) which is capable of braking at 0.5g. With this autonomous system running on-board, ``gotthard driverless'' can cruise at a top speed of 54kmph.

\begin{figure}[h]
    \centering
    \includegraphics[width=0.8\textwidth]{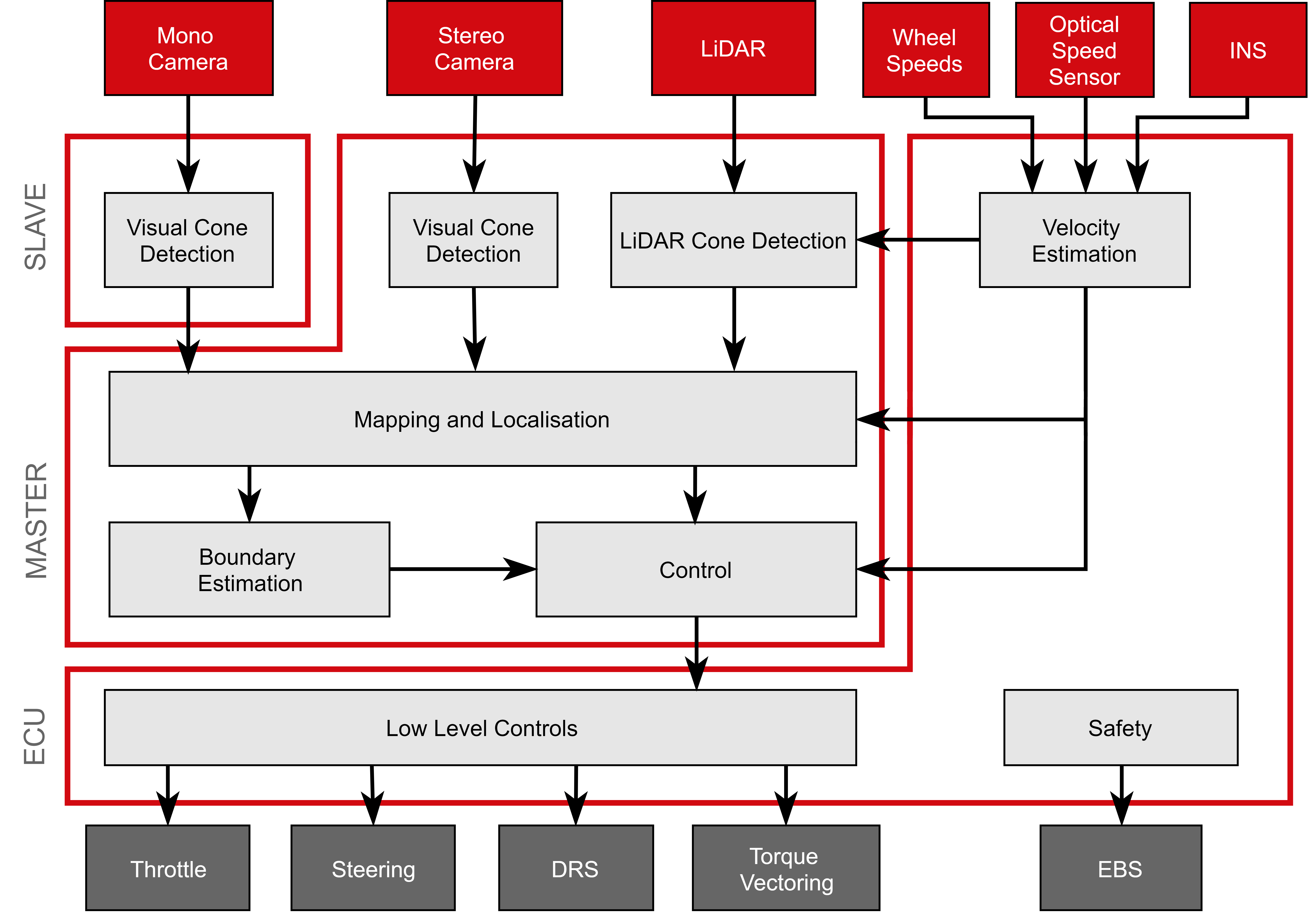}
    \caption{The pipeline running real-time on ``gotthard driverless'' giving it autonomous driving capabilities. Most modules run on the rugged PIP-39 running an i7 with an Nvidia GTX 1050i. The cameras' raw sensor data as well as the monocular pipeline is handled by the Jetson TX2 which is slave-configuration to the PIP. See Figure \ref{fig:computing_system} for details on the computing system.}
    \label{fig:system_arch}
\end{figure}

\begin{figure}[h]
    \centering
    \includegraphics[width=1.0\textwidth]{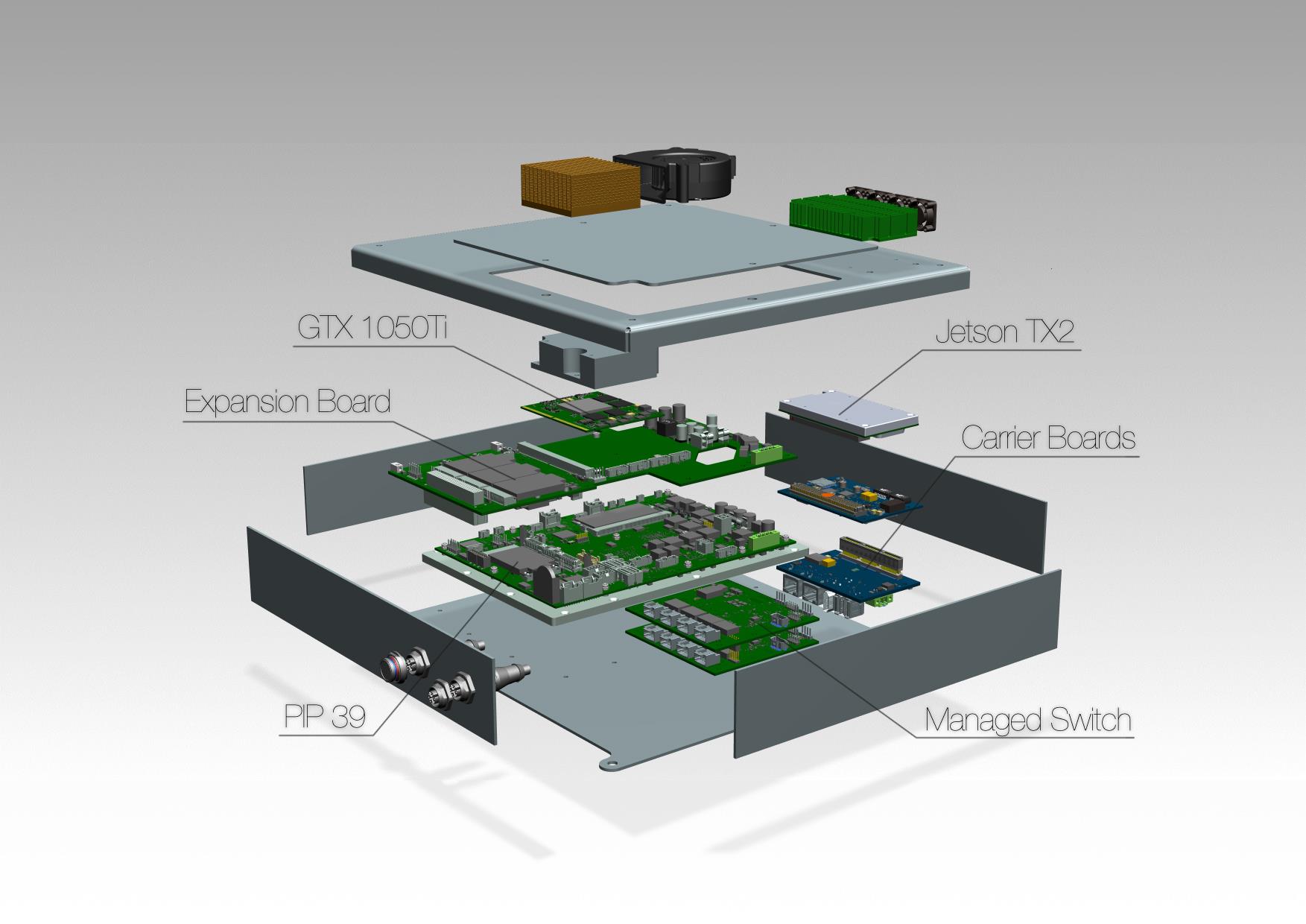}
    \caption{View of the customized, high-performance, low-power computing system on ``gotthard driverless" to run the autonomous system stack. The communication across devices and computers is handled through ethernet via switches. The low-level communication is handled using CAN lines.}
    \label{fig:computing_system}
\end{figure}

To be able to process raw data and pass it through various modules, and to obtain the throttle command and the steering angle, a powerful computing system with real-time processing capabilities is required. On ``gotthard driverless'' the autonomous system runs on an efficient, low-power, customized computing system. It comprises of a rugged, PIP-39 running an i7 with an Nvidia GTX 1050i GPGPU and an Nvidia Jetson TX2 running in a slave configuration to the PIP. The cameras are directly connected to the slave TX2 and it is also where part of the computer vision functionality is processed (the monocular pipeline, which is the focus of this work). The PIP-39 runs the rest of the modules such as LiDAR, SLAM, boundary estimation and control. A low-level electronic control unit (ECU) estimates the state of the car and manages state transitions on the car's autonomous state machine.

\section{``Trackdrive" - The Need for Computer Vision}

There are two important ingredients to complete such events, (1) a reliable hardware platform, which would be ``gotthard" in our case and (2) a robust autonomous system pipeline that works in all possible conditions and circumstances. We build on top of the electric car prototype ``gotthard" built in 2016 by AMZ and revamp it successfully to convert it into ``gotthard driverless" by adapting its design and the autonomous pipeline to breathe life into ``gotthard''.

As mentioned earlier, the ``Trackdrive" tests the capabilities of the autonomous system and is the most relevant to us. To drive on an unknown track, the car needs to be capable of reliably and accurately perceiving the environment around it. As per the rules of the competition, the track is marked by cones. The left and right track limits are marked by blue and yellow traffic cones respectively. Obviously, in broader use-cases, one would need to perceive a variety of agents both animate and inanimate from moving humans to stationary vehicles among a variety of other objects. But due to the specific nature of this scenario, these very cones are the landmarks that we are most concerned about. We design a system that perceives cones, accumulates measurements over time and puts them on a common map to finally be able to use it to extract feasible trajectories for the car and execute throttle and steering commands to the actuators.

We derive the goals of the computer vision sub-system from the team goals we concluded on at the beginning of the season which were as follows:
\begin{itemize}
    \itemsep-0.5em
    \item Maximise points in a Formula Student Driverless event
    \item Finish all disciplines (dynamic events)
    \item Win Formula Student Germany
    \item Safety is a priority
    \item Lay down a solid foundation for AMZ Driverless
\end{itemize}

In order to achieve these team goals it was concluded that a computer vision perception module is necessary. There were several reasons to include it in our autonomous system architecture stack:
\begin{itemize}
    \itemsep0em
    \item \textbf{Need for perception to map and localize}: Making the map with cones as landmarks is key to driving at higher speeds without accruing penalties by hitting cones. Once the map is frozen after the first lap (loop closure), perception is still required to accurately localize the robot's position within this map.
    \item \textbf{Accurate color information}: With color information, the ambiguity of possible trajectories is drastically reduced and it also provides a means to reduce the search space. This allows to resort to quicker, more efficient and effective heuristics to ascertain the driving line.
    \item \textbf{Good position estimates and a redundant perception system}: A LiDAR has precise and accurate position estimates but the number of point returns reduces drastically with increase in distance (as the beams diverge). With a computer vision pipeline that provides accurate position estimates with color information, one can practically map and localize only with cameras in case the LiDAR pipeline fails.
    \item \textbf{Extend range of perception to help anticipate better}: If the system can accurately perceive cones for a long range of distance, this allows the car to go faster. More cones implies path planning and generating a control sequence can be performed for a longer horizon.
\end{itemize}

The goals of the computer vision module that we defined for ourselves were as follows:
\begin{itemize}
    \itemsep-0.5em
    \setlength{\itemindent}{1cm}
    \item Provide, for each cone detected by the pipeline,
    \vspace{-0.3cm}
        \begin{itemize}
            \itemsep-0.5em
            \setlength{\itemindent}{1cm}
            \item \textbf{Probability}: $P(yellow cone)$, $P(blue cone)$ and $P(orange cone)$
            \item \textbf{Pose}: Position of cone in 3D
        \end{itemize}
    \item Avoid false positives at all costs i.e. misclassifications
    \item Computationally efficient
    \item Accurate position estimates
    \item Modular and an easy-to-debug pipeline with interchangeable sub-modules
\end{itemize}

\section{Focus of this Work}
As introduced earlier, ``gotthard driverless'' features 3 cameras for perceiving its surrounding. Two of which act as a stereo pair and the third as a stand-alone monocular camera. This work focuses on the monocular pipeline and how a priori information about objects of interest is exploited to detect and estimate 3D cone positions accurately up to 15 meters using just a single image from the monocular camera. A novel feature regression scheme, ``keypoint regression'' is introduced which is used to match 2D-3D correspondences. The work shows how 3D priors about the object of interest can be incorporated in the form of a novel \textit{cross-ratio} term in the loss function and also use the same 3D priors to accurately estimate 3D pose for multiple objects with a single image, which typically would have been an ill-posed problem.

The thesis not only presents what can be done in theory but shows the practical potential and application of the proposed low-latency, lightweight, pipeline on ``gotthard driverless'' to perceive the surrounding environment, help create an accurate map and drive autonomously, without human intervention, at high speeds.

\section{Thesis Organization}
The thesis discusses how the aforementioned goals for the computer vision module were achieved in terms of the monocular pipeline. We skim over the design choices of hardware, sensors and their positioning on ``gotthard driverless''. Then turn focus to the monocular pipeline and discuss in detail the approach and its working. Each sub-module of the pipeline is discussed in detail. Finally, discuss and present the results of the pipeline followed by analysis and possible improvements in the form of future work.

% Give an introduction to the topic you have worked on:

% \begin{itemize}
%  \item \textit{What is the rationale for your work?} Give a sufficient description of the problem, e.g. with a general description of the problem setting, narrowing down to the particular problem you have been working on in your thesis. Allow the reader to understand the problem setting. 
%  \item \textit{What is the scope of your work?} Given the above background, state briefly the focus of the work, what and how you did.
%  \item \textit{How is your thesis organized?} It helps the reader to pick the interesting points by providing a small text or graph which outlines the organization of the thesis. The structure given in this document shows how the general structuring shall look like. However, you may fuse chapters or change their names according to the requirements of your thesis.
% \end{itemize}

%% ----------------------------------------------------------------------------
% BIWI SA/MA thesis template
%
% Created 09/29/2006 by Andreas Ess
% Extended 13/02/2009 by Jan Lesniak - jlesniak@vision.ee.ethz.ch
%% ----------------------------------------------------------------------------
\newpage
\chapter{Related Work}
Autonomous driving has become one of the most ambitious as well as most prized problem to be tackled jointly by the computer vision, robotics and machine learning community. Due to its interdisciplinary nature, it is imperative for diverse modules such as controls, perception, mapping and actuators to work in tandem to achieve the goal of driverless vehicles.

The DARPA Grand Challenge\cite{darpaGC} marked the beginning of a long, arduous journey among the research community to realize this goal of self-driving vehicles and transform it from a blueprint in the labs to a reality on the road.

Industrial and large-scale projects have been undertaken by companies such as Uber, Tesla, Google's Waymo, nuTonomy and German automotive giants Mercedes, BMW and Audi to name just a few.

To further accelerate this, in 2017, Formula Student Germany, the biggest student engineering competition in the world introduced a new category apart from the pre-existing \textit{combustion} and \textit{electric} categories, the \textit{driverless} class. The idea is to encourage development of such vehicles from student only teams. It takes about 9 months to design a concept and to build an autonomous race-car; hardware, software and everything in between. Last, year's driverless car from AMZ ``fl{\"u}ela driverless'' was the winner of the first driverless event in Germany, 2017. The car used a stereo camera and performed tracking of cones with the help of ORB features \cite{fluela, mur2015orb}. The range of the system was short and cones only a few meters from the car could be perceived.

Object detection is one of the sub-modules of the monocular pipeline and ``gotthard driverless'', AMZ's driveless car for the 2018 season. One of the most prominent and well-known schemes was the Viola-Jones' lightening fast face-detector \cite{viola2001rapid}. It employed weak learners to accurately detect faces using Haar-based features. Dalal and Triggs \cite{dalal2005histograms} tried to address pedestrian detection by coupling a support vector machine (SVM) with histogram of oriented gradients or HoG to localize walking humans in images. Moving forward, a deformable parts model approach that used convolution with feature maps was developed by Felzenszwalb et al. \cite{felzenszwalb2010object}. The idea was to find regions of highest aggregated response for different parts of the object of interest, eventually using this to output detections.

The next class of well-known object detectors used deep learning in the form of convolutional neural networks (CNNs). The string of R-CNN \cite{girshick2014rich, ren2015faster, girshick2015fast} schemes used CNN-based features on region proposals.

YOLO: You Only Look Once \cite{redmon2016you} shot to fame and became a household name in computer vision because of its ingenious concept of detecting multiple objects with a single pass of the neural network. Doing this accurately and at a high frequency was the break-through making a multitude of real-time applications possible. By cleverly using regression to classify and detect objects, YOLO became very popular with the community. Liu et al.'s SSD or Single shot detector \cite{liu2016ssd} is another work that had a similar flavor to YOLO where objects are detected in a single pass through the convolutional neural network. ``gotthard driverless'' utilizes YOLOv2 \cite{YOLOv2} as the object detector in the pipeline.

One of the main contributions of this work is to be able to accurately estimate object pose upto 15 meters using just a single frame. A priori information about the 3D geometry is used to regress highly specific feature points called \textit{keypoints}. Previously, pose estimation and keypoints have also been tackled by other works. \cite{savarese20073d, glasner2012aware}. Glasner et al. \cite{glasner2012aware} estimate pose for images of cars using an ensemble of voting support vector machines (SVM).

Tulsiani et al. use features and convolutional neural networks to predict view-points of different objects \cite{vp_and_kp}. Their work tries to capture the interplay between viewpoints of objects and keypoints on those specific object classes.

PoseCNN directly outputs the 6 degrees-of-freedom pose using deep learning \cite{posecnn}. Gkioxar et al., \cite{gkioxari2014using} use a k-part deformable parts model and presents a unified approach for detection and keypoint extraction on people.

Our work differs from previous works by not only estimating long distance position of objects but also presenting in-depth analysis of the various sub-modules in the pipeline. In addition, we introduce a novel ``keypoint regression'' scheme with a \textit{cross-ratio} loss term to exploit a priori 3D information about the object geometry and appearance. Finally, this work is implemented on the computing system on ``gotthard driverless'' to compete in Formula Student Driverless 2018 competitions held in Germany and Italy. The system runs efficiently and in real-time, providing accurate object position estimates up to 15 meters on an autonomous race-car, allowing it to create an accurate map and cruise at a top-speed of 54 kmph.

Finding feature points of interests, also known as keypoints, has been studied and researched extensively in the context of human pose.

In the past there has been work to improve the perspective n-point algorithm. P3P works with exactly 3 points which would require discarding more half of our correspondences and many works have proposed a closed-form solution for the same. EPnP strives to make it computationally efficient with an algorithm that is linear in number of 2D-3D correspondences \cite{lepetit2009epnp}. DLT or direct linear transform is another method to estimate pose using 2D-3D correspondences and does so by solving a system of equations. But it is less stable with more points and also has degeneracies for planar points. In this work we use the iterative non-linear PnP using non-linear least squares \cite{opencv3Drecon}. As there are only 7 points and it is quick to compute. It is further used with Random Sample Consensus or RANSAC to be resilient to outliers.

% Describe the other's work in the field, with the following purposes in mind: 

% \begin{itemize}
%  \item \textit{Is the overview concise?} Give an overview of the most relevant work to the needed extent. Make sure the reader can understand your work without referring to other literature.
%  \item \textit{Does the compilation of work help to define the ``niche'' you are working in?} Another purpose of this section is to lay the groundwork for showing that you did significant work. The selection and presentation of the related work should enable you to name the implications, differences and similarities sufficiently in the ``discussion'' section.
% \end{itemize}

%% ----------------------------------------------------------------------------
% BIWI SA/MA thesis template
%
% Created 09/29/2006 by Andreas Ess
% Extended 13/02/2009 by Jan Lesniak - jlesniak@vision.ee.ethz.ch
%% ----------------------------------------------------------------------------
\newpage
\chapter{Sensor Suite Concept and Design}

\begin{figure}[h]
    \centering
    \includegraphics[width=1.0\textwidth]{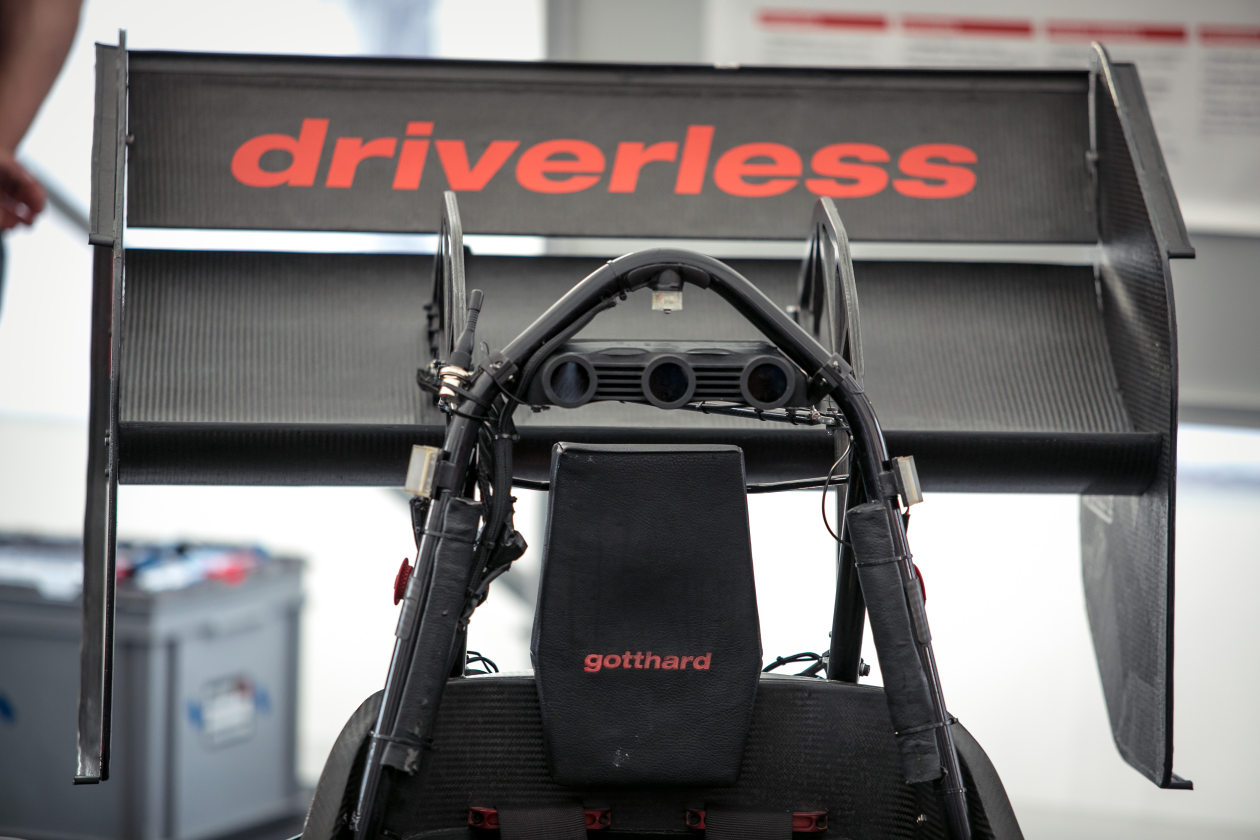}
    \caption{Camera housing on ``gotthard driverless'' at Formula Student Germany 2018, Hockenheimring. Image credits: Schulz, FSG.}
    \label{fig:camera_FSG}
\end{figure}

Setting up a perception pipeline on a real-time system involves a coherent and clever interplay between software and hardware. Although, undermined more often than not, choosing appropriate hardware can greatly benefit the whole system. With proper choices it can improve raw sensor data flow management, sub-system performance and eventually the overall throughput of the system.

To ensure an efficient and effective perception system we build a customized camera hardware setup with choices based on goals and requirements defined at the onset. We postpone the discussion of the novel ``keypoint regression'' scheme and consequently position estimates of multiple objects from a single image by exploiting 3D object priors until the next chapter. The main focus of this chapter is to discuss the development of the hardware setup and raw data acquisition which is as crucial, if not more, as the software and processing of raw data.

\section{Design Choices}
As mentioned earlier, choosing the most relevant sensor based on the application can already have a monumental impact on the pipeline. Keeping multiple constraints such as limited on-board computation and sharing with other modules, sensor envelope (which defines restrictions on where sensors can be placed on the car according to Formula Student regulations) and the package goals. We achieve a redundant pipeline by having a LiDAR and cameras function and provide cone position and color independently. This allows ``gotthard driverless'' to still be able to run in case of single mode failure.

\subsection{Choosing the right sensor}

The computer vision pipeline has a 3 camera setup. Two cameras, on the extremities, act as a stereo pair to help triangulate cones close to the car from 2 meters up to 8 meters away from the car. The center camera is a stand-alone monocular camera which has a range that allows it to perceive cone positions accurately up to 15 meters. The mono camera runs the novel pipeline which is the main focus of this work.

\begin{itemize}
    \itemsep0.0em
    \item \textbf{Noise and corruption free transmission}: After feedback from previous season, it was noticed that there was noise in the sensor raw data due to electro-magnetic interference (EMI) due to the various electric components as well as high-voltage that drives the motors. This was primarily due to the USB transmission of data to the computers from the cameras. To prevent this from happening, we choose cameras with GigE (Gigabit Ethernet) capabilities for a noise-free transmission.
    \item \textbf{Neutralizing sudden changes in lighting}: To ensure that glares and sudden lighting changes do not render the cameras useless, CMOS (Complementary Metal-Oxide Semiconductor) sensors are chosen over CCD (Charged Coupled Device). CCD sensors are susceptible to washed out image pixels due to glare while CMOS based sensors are not. Additionally, the camera housing has polarized filters mounted in front of each of the camera optics.
    \item \textbf{Capturing images at high speeds}: Since, ``gotthard driverless'' will be cruising at relatively very high speeds as compared to your average robotic system, it is crucial that the captured images are not distorted due to high-speed motion. This would change the appearance of the physical scene as captured on the image, making 3D measurements from image quite inaccurate. Since, we want to estimate 3D position of multiple objects this affect is undesirable. Cameras are generally manufactured with a rolling shutter fundamentally due to their cost-effectiveness. Rolling shutter cameras read out the image in a scan-line fashion causing distortions in case of high speed operation. We use global shutter cameras instead to counter such distortions in the image.
    \item \textbf{Dramatically increase range of computer vision perception system}: By cleverly choosing the optics we are able to have a combined range of up to 15 meters from the complete computer vision perception system. The stereo is meant to accurately triangulate nearby cones and has a shorter focal length (5.5mm). While the monocular camera, by using prior information about 3D object's shape, size and geometry to estimate depth for cones that are far away, is equipped with a lens  of a longer focal length acting like a "magnifier" (but with reduced field of view). By smartly choosing the optics, we are able to almost double the range the cameras can see, which is 5-6 meters more than limit of the LiDAR (VLP-16 Hi-res) pipeline estimating cone positions.
\end{itemize}

\subsection{Potential options that were considered}

Other setups and hardware equipment was considered but eventually, the above mentioned concept was the most relevant one. Some options that we chose not to go forward with as part of our design are listed below with justifications for the same.

\begin{itemize}
    \itemsep0.0em
    \item Commercially available (stereo) cameras generally have a shorter baseline (while we could accommodate for more on the roll hoop). Some of the customized configurations are heavy (about 5 kilograms) and have additional computation units packaged into one, which may not be the best choice for putting on a race-car. These off-the-shelf systems also have fixed focal length for the cameras, which is generally very short and may not allow to perceive and estimate more than 5-6 meters away from the camera.
    \item Some companies do offer to build special, on-order customized solutions according to given requirements and specification however these are extremely expensive as compared to building a customized setup from scratch.
    \item Since, we had decided on the concept of having 3 cameras, the ones on the extremes as a stereo pair while the central one as a monocular camera, there were no such commercially available configurations that also managed to not violate the sensor envelope and Formula Student regulations.
    \item Obviously, another design choice would be to have more than 3 cameras with a different concept. However, there is a trade-off between more information by having more cameras on ``gotthard driverless'' and the amount of data that needs to be transmitted and received, its overhead and most crucially, the on-board computation power to process such amounts of data. 3 cameras running at 10 frames per second with 2 megapixels images already transmit about 1 Gigabit of data per second! Having more cameras would compel revisiting the computing system design and data transmission cables on the car.
    \item We considered off-the-shelf cameras from Point Grey (Bumblebee stereo), Carnegie Robotics and Stereo Labs (ZED stereo camera). All of which had one or more drawbacks mentioned above.
\end{itemize}

With the above options and their downsides, it was concluded to develop a custom hardware setup for the computer vision perception pipeline.

% commercially available base lines are short, heavy, no custom focal length
% customized solutions offered by companies are expensive
% 3 camera setup, fitting on the roll hoop without violating sensor envelope
% off-the-shelf stereo have USB cables, cutting them is not recommended
% bumblebee, basler, point grey, CR, ZED
% trade off between number of cameras, data transmission and computing system requirements

%https://robotics.stackexchange.com/questions/896/how-to-select-cameras-for-a-stereo-vision-system/938#938
%https://robotics.stackexchange.com/questions/11749/estimate-disparity-error-for-depth-accuracy-estimation

\section{Camera Housing}

\begin{figure}[h]
    \centering
    \includegraphics[width=0.7\textwidth]{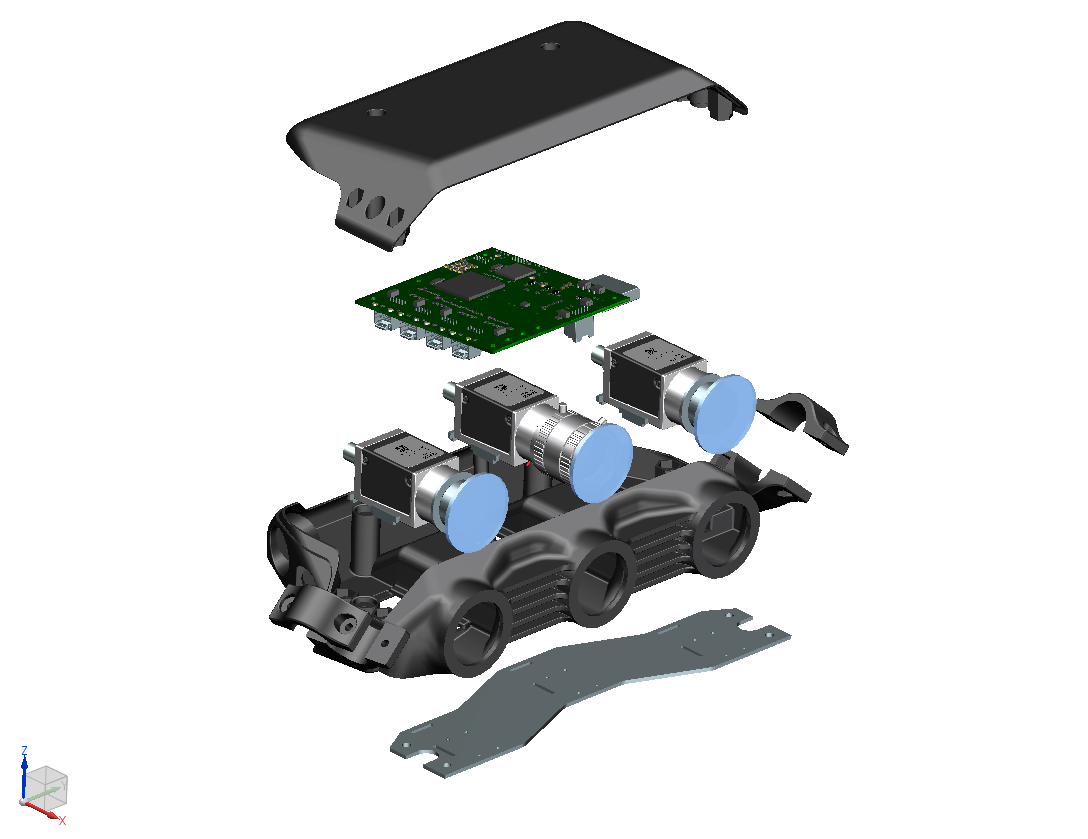}
    \caption{``gotthard driverless" computer vision perception system in a self-designed, waterproof 3D printed housing with passive cooling. camera housing.}
    \label{fig:camera_housing}
\end{figure}

\subsection{Self-designed, 3D printed shell}
The cameras are housed in a self-designed, custom 3D printed shell. The housing itself has 3 main parts, a main body, a top cover and a bottom plate made out of aluminum. The 3 cameras are mounted directly on the aluminum plate with inserts to aid proper positioning. Each of the Basler ``ace'' cameras have two cables, one for power and triggering and another Ethernet cable for data transmission.

\subsection{A neat data transmission solution}

In order to have a neat design, we use a rugged switch, mounted above the cameras, that multiplexes the 3 data transmission Ethernet cables to a single output Ethernet carrying raw sensor data from all 3 cameras. This allows the housing to have only two connectors, one for powering the electronics inside the shell and another for transmission of data through a single Ethernet cable.

\subsection{All weather camera operation}
On a hot summer's day, such as the competition week during FS Italy 2018, 3 cameras and a switch in an enclosed housing can heat up rapidly. In order to keep the the cameras at optimal operating temperature, we use the aluminum plate as a passive cooling source, where the cameras are physically in contact, allowing heat transfer out of the housing shell. The housing is also sealed and water-proof to have smoothly functioning hardware even in torrential conditions.

The left and right cameras (stereo pair) have 5.5mm short focal length optics, while the center camera (stand-alone monocular camera) has a longer 12mm focal length. To ensure ideal raw images, we use polarized filters which are also hydrophobic, ensuring best raw image quality even under influence of potential glares or water droplets.

% Discuss the internals. Ethernet, cooling, water proof, switch, lens, filters.

\section{Sensor Positioning}
% Discuss good vantage point and sensor envelope. Also discuss potential positions that were considered. Keep the 2 sensors physically away.

\begin{figure}[h]
    \centering
    \includegraphics[width=1.0\textwidth]{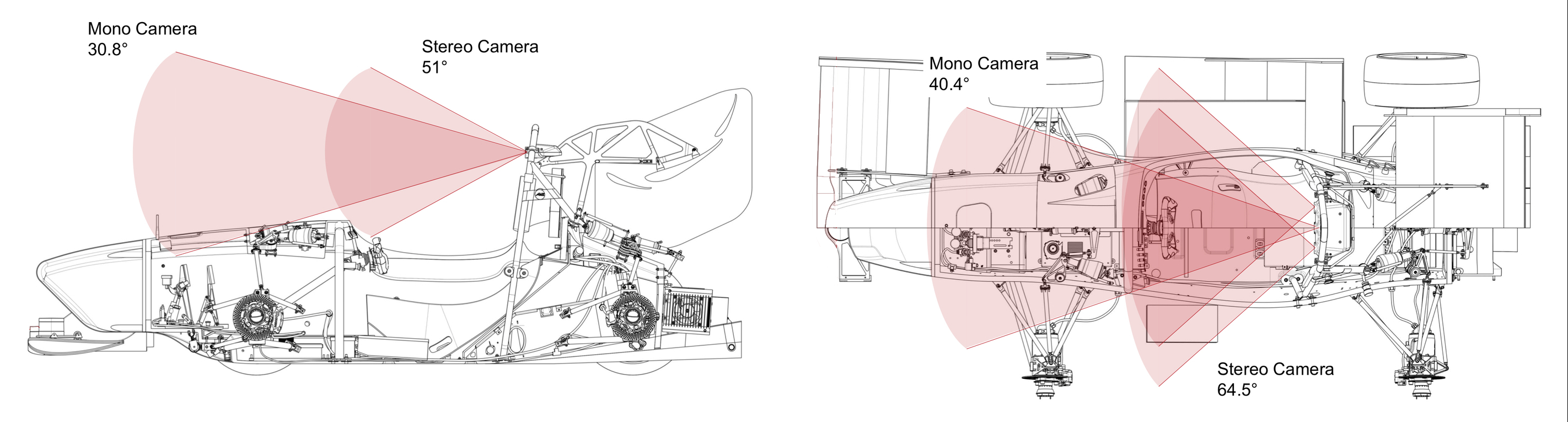}
    \caption{``gotthard driverless" cross-section and top-view. The cameras are all mounted on the roll-hoop. The cameras on the extreme act as a stereo pair with a 5.5mm focal length optics while the center camera is a stand-alone monocular camera with a longer, 12mm focal length lens.}
    \label{fig:gotthard_w_cam}
\end{figure}

The camera housing has the provision to be mounted on to the roll hoop with clamps from two sides. As can be seen in Figure \ref{fig:gotthard_w_cam}, the cameras are placed above the cockpit. This is the highest possible positioning without violating Formula Student regulations. The stereo pair with a shorter focal length (5.5mm) have a larger field-of-view (FOV) $51^\circ \times 64.5^\circ$ . The monocular camera, on the other hand, due to its longer focal length (12mm) has a shorter field-of-view of $30.8^\circ \times 40.4^\circ$. The stereo pair is calibrated for a range between 2 meters to 8 meters. The monocular camera is calibrated for a longer range, 4 meters to 15 meters. To calibrate the intrinsics and extrinsic for the cameras over such large distances, commonly used A4-paper sized checkerboard are not sufficient as they are only detected up to a distance of less than 1 meter from the image plane. A 210cm x 167cm checkerboard was built to calibrate the computer vision perception system's stereo and monocular cameras for ``gotthard driverless''

During the design process, placing the cameras at other plausible locations such as under the nose (on either side of the LiDAR) were considered. But this poses other issues. Having the two sensors at the same physical location on ``gotthard'' makes it non-redundant in case of a crash. If the front wing is damaged, one would lose perception capabilities from LiDAR as well as camera due to the direct damage to the sensors. A bigger consequence of placing sensor would be the dramatic increase in overlapping cones seen in the images. This may seem trivial at first but could have a dramatic affect on the performance of the computer vision perception pipeline. This would have far-reaching effects, making it more challenging to detect cones in general and eventually giving the computer vision pipeline a hard time estimating position of these cones most of which are occluding and overlapping each other. Placing the cameras at the current location allows the best vantage point, over looking the track with minimal cone overlap.

\section{Image Acquisition}
% Discuss FPS, triggering, cropping, transmission, connection to TX2. Show exemplary images from stereo and mono.

\begin{figure}[h]
    \centering
    \includegraphics[width=1.0\textwidth]{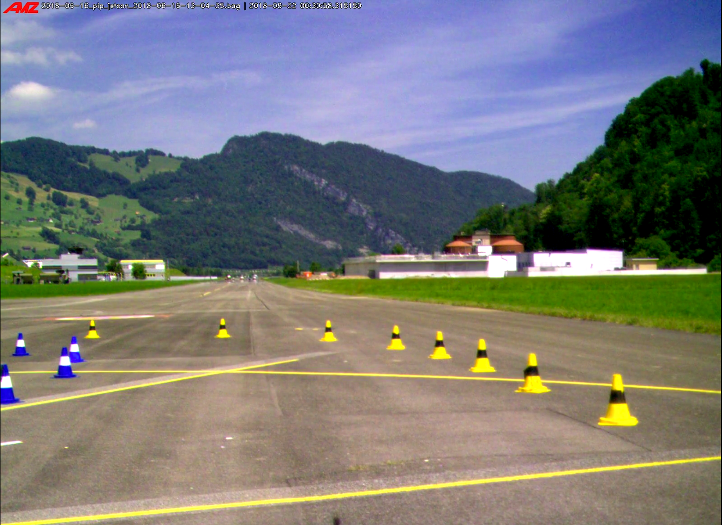}
    \caption{View from the monocular camera on ``gotthard driverless''.}
    \label{fig:mono_image}
\end{figure}

\begin{figure}
	\centering     %%% not \center
	\subfigure[Left camera]{\label{fig:a}\includegraphics[width=0.48\textwidth]{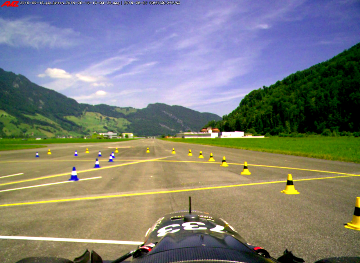}}
	\subfigure[Right camera]{\label{fig:b}\includegraphics[width=0.48\textwidth]{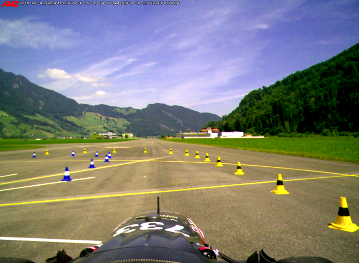}}
	\caption{View from the stereo pair setup on ``gotthard driverless'' of a track laid out from yellow and blue cones.}
    \label{fig:stereo_image}
\end{figure}

The cameras on ``gotthard driverless'' acquire raw images at the rate of 10 frames per second. To ensure the images are captured at the same instance of time, all three cameras receive a clock signal to trigger the frame capture simultaneously. As mentioned above, the cameras output raw data is sent over through a single Ethernet cable directly into one of two Ethernet ports on the Jetson TX2. The other Ethernet port of the Jetson TX2 is used to establish a connection to the master computer, the rugged PIP-39.  Exemplary raw images from the sensor can be seen in Figure \ref{fig:mono_image} for the monocular camera and Figure \ref{fig:stereo_image} for the stereo pair.

Since, in our specific scenario the landmarks that truly matter are cones that form the track, upper part of the image is cropped on the sensor side itself i.e. a $1200 \times 1600 \times 3$ image is cropped to a $800 \times 1600 \times 3$ by truncating the top 400 pixels. By eye-balling the data one can safely assume that the top 400 pixels correspond to the sky in the image. This reduces the futile data overhead and ensure efficient data transfer.

% The objectives of the ``Materials and Methods'' section are the following: 
% \begin{itemize}
%  \item \textit{What are tools and methods you used?} Introduce the environment, in which your work has taken place - this can be a software package, a device or a system description. Make sure sufficiently detailed descriptions of the algorithms and concepts (e.g. math) you used shall be placed here.
%  \item \textit{What is your work?} Describe (perhaps in a separate chapter) the key component of your work, e.g. an algorithm or software framework you have developed.
% \end{itemize}

\chapter{Monocular Camera Perception Pipeline}
% \textbf{
% Using DL/ML where necessary (for feature extraction and exploit data) and classic computer vision where possible. Direct feature extraction. Exploiting prior information
% Increase range and reduce computation resources required light-weight and efficient pipeline}
This section shifts the focus on how to estimate 3D position of multiple objects from a single image. Although, it is an ill-posed problem but with a priori information in the form of the shape, size and geometry of the object-of-interest, this is solvable, as elaborated in this chapter.

With the rise of deep learning, the availability of tremendous amounts of data and extremely powerful hardware (GPGPUs) there has been an explosion of deep learning papers, especially ones showcasing the power of end-to-end learning. Although, on paper it seems very fancy and sophisticated, wielding such powerful black boxes is no child's play. Diverse data is required to ensure that such models do not over-fit to data specific scenarios and fail miserably when unseen data is thrown at them.

One might be tempted to attempt an end-to-end image to pose estimation or even image to throttle and steering command \cite{nvidiaBojarskiTDFFGJM16}. However, we employ a part-based approach, making sure we can run on-board in real-time reliably and have the power of interpreting data and sub-module outputs. This allows for easy debugging and extensive testing, indirectly helping to optimize each sub-module until a certain level of performance is reached by the pipeline as a whole system. Also, a part-based approach allows for easy replacement of sub-modules with better performing or more efficient counterparts as desirable.

The pipeline's sub-modules are run as nodes using Robot Operating System or ROS \cite{ros} as the framework that eases handling of communication and data messages across multiple systems as well as different nodes. Different sub-modules communicate via messages, they receive data and output processed information. Another important aspect is that ROS is open-source and provides tools for visualization, monitoring and simulation, making it easy to integrate, test, diagnose and develop the complete software system.

The idea is to exploit data where it makes most sense, such as extraction of very specific features, while use results from classical computer vision and mathematics to ensure robust, reliable estimates. This prevents from using deep learning as an overkill solution to problems that already have well-established, mathematical solutions.

As mentioned in earlier chapters, there are two parts to the computer vision perception system: the stereo and the monocular pipeline. The stereo pipeline use the sub-modules explained in this section to have an extremely efficient way of triangulating and estimating depth from binocular vision. This methodology of drastically reducing the search space and cleverly tackling the issue of having numerous and often incorrect feature matches, discussed briefly in a later section. For now, the focus is on elaborating how 3D pose estimation was performed on ``gotthard driverless'' from the monocular pipeline on a per-frame basis.

\section{Pipeline Overview}
The monocular pipeline has 3 crucial sub-modules which enable it to detect multiple objects of interest and accurately estimate their 3D position up to a distance of 15 meters by making use of a single measurement in the form of an image captured by the monocular camera.

\begin{figure}[h]
    \centering
    \includegraphics[width=1.0\textwidth]{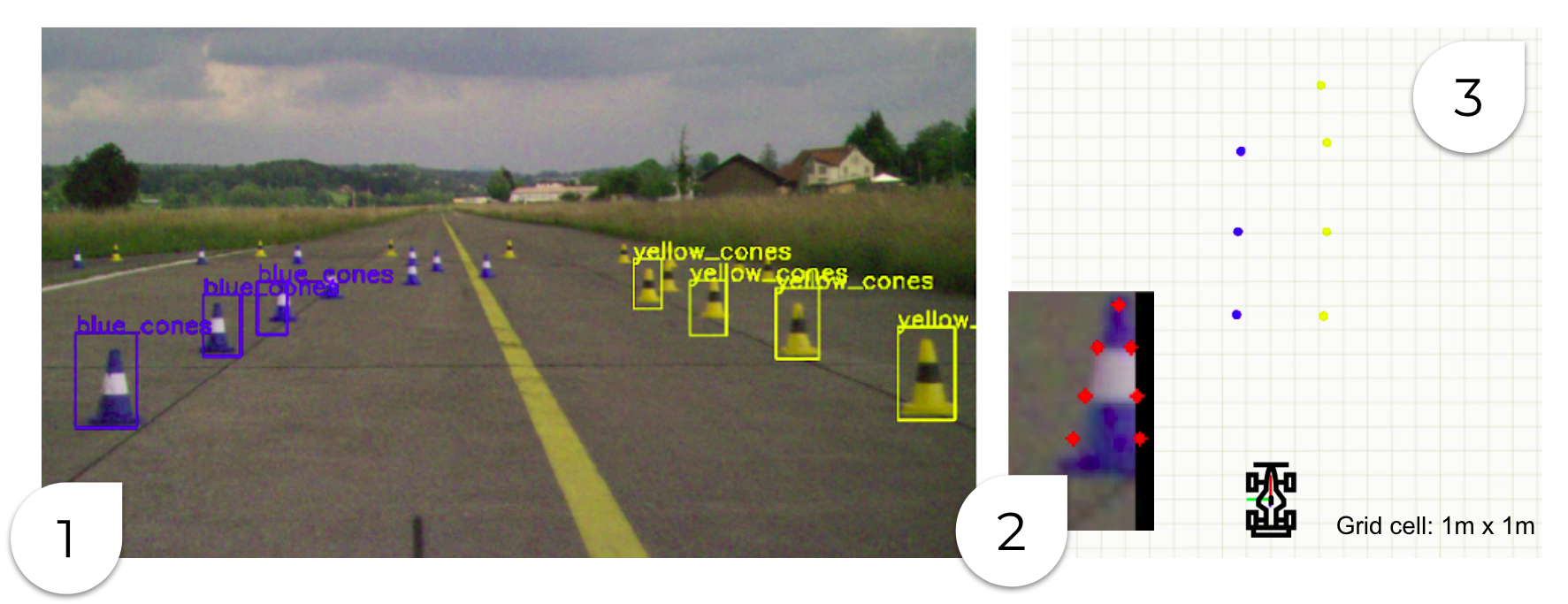}
    \caption{The monocular pipeline can be broken down into three parts. (1) Multiple object detection, (2) Keypoint regression and (3) 2D-3D correspondence followed by 3D pose estimation from a single image. The three sub-modules are depicted in this pipeline overview.}
    \label{fig:pipeline_overview}
\end{figure}

The following sections discuss each of the sub-modules in detail with the idea in mind behind having these in the first place. Since, during FS events, the weather may be unpredictable, it is of utmost importance to have a very robust pipeline. A part-based approach can only be as robust and reliable as its weakest sub-module. As mentioned earlier, having the right hardware which works in calamitous and extreme conditions such as rain or hot summer days, that is only one part of the story. Not just the hardware but also the software needs to be equally reliable. The following sections also discuss how the sub-modules are made to be robust and work reliably in changing weather and diverse lighting conditions.

\section{Multiple Object Detection}
Before one can actually estimate 3D position of multiple objects from a single image, it is necessary first to be able to detect these objects of interest given a single image. 

Object recognition has 4 main categorizes of tasks: (1) classification, (2) classification and localization, (3) object detection and (4) instance segmentation. The first two correspond to having a single object while the latter cover multiple objects in a single image.

Before the advent of GPGPUs and deep learning based approaches, classical computer vision was used to detect objects given an image.

Instead of using slow and computationally intensive cascade and sliding window approaches, we employ a quick, real-time and powerful object detector in our pipeline in the form of YOLOv2. Its ability to be fine-tuned with lesser data (using the ImageNet dataset \cite{imagenet}) pre-trained weights and robust outputs made it the right fit in out system.

\subsection{Importance of color information}
Last season, rules permitted, mapping cone locations via a track-walk device, such as a pip-point accurate GNSS system. This allowed teams to have a prior about the shape of the track and how the cones were arranged. This made it relatively simple as it was more about localizing correctly and finding the correct cone correspondences received from the perception on the car and the obtained map. However, this season, the rules prohibit use of any such device and state that no external devices, except the car itself can be used. This makes it very crucial to not only detect cones but also what color they are. Without having a pre-mapped track and color information, the path planning module faces a lot of feasible paths which are incorrect and will drive the car out of the track. This could be paths that let the car pass through cones of the same color or treating noisy returns from the LiDAR as cones (such as tall grass which has similar characteristics as the volume of a point cloud containing a cone).

\begin{figure}[h]
    \centering
    \includegraphics[width=1.0\textwidth]{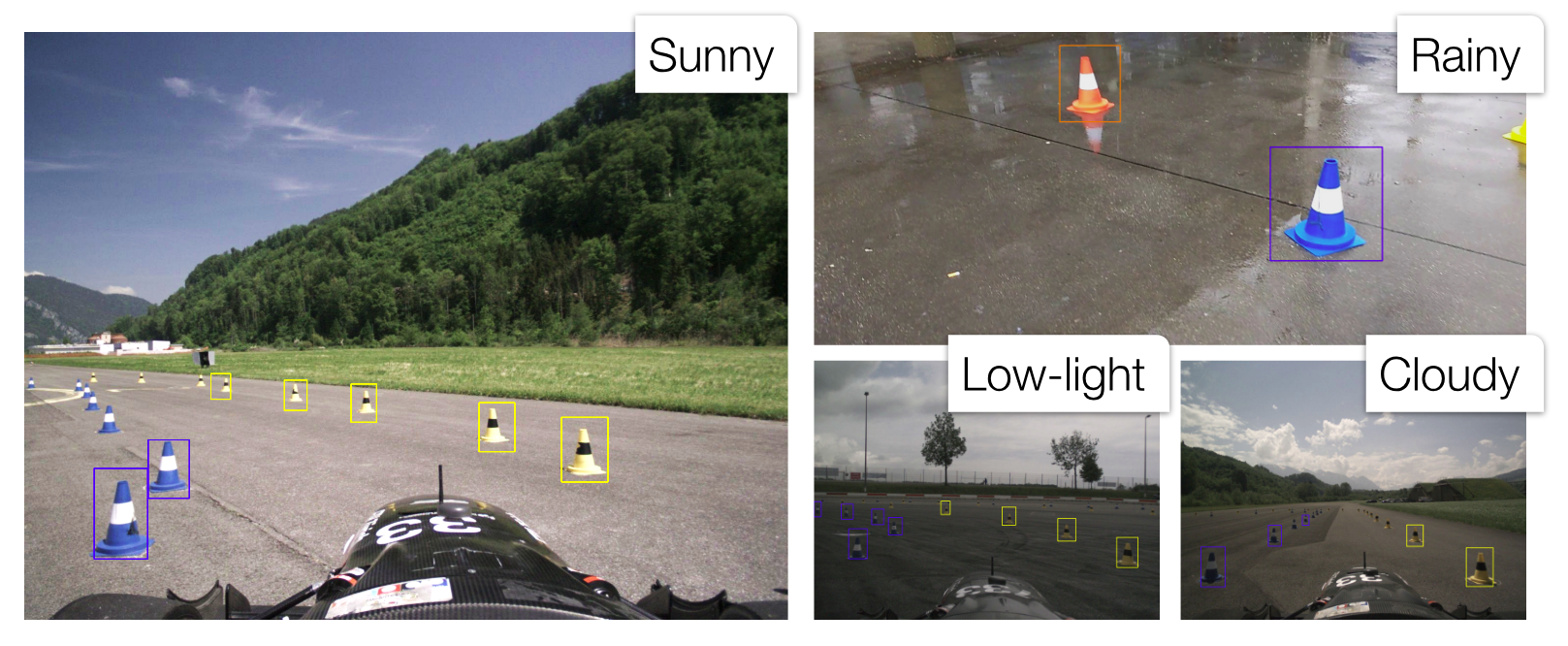}
    \caption{Exemplary images from varying lighting and weather conditions showing robust performance from the object detection to detect `yellow', `blue' and `orange' cones. It is imperative to have a robust cone detection as it is the first sub-module in the pipeline and directly affects performance of the sub-modules that follow, eventually the final output as well.}
    \label{fig:robust_detection}
\end{figure}

To tackle and prevent such scenarios, the computer vision perception system had it as a goal to also detect colors of cones using information from the image. The path planning then has a cost function with a penalization term for potential paths that drive the car through same colored cones. This gives the car a high chance of completing the first lap where the car has the highest chance of going off-track as the track is unknown.

One could detect color by naively comparing pixel values and evaluating whether a patch of pixels is more yellow or more blue. But due to frequently changing illumination and weather as well as broken cones, this may not be very robust. We design the detector such that the cone color information can be directly obtained from it. In other words, we treat each colored cone as a different class for the object detector.

As mentioned before, color is the most important and reliable piece of information from the camera sensors that keeps the car on track and prevents it from going out of track limits. SLAM makes a map with cones as landmarks and then path planning is done by extracting boundaries that serves as left and right track limits (demarcated by blue and yellow cones). We show exemplary images in diverse weather and lighting conditions and show the robustness of this particular object detection module, customized to detect colored cones.

\subsection{Customizing YOLOv2 for Formula Student Driverless}

The object detector should be efficient in that it is fast, requires lesser memory and is still decently accurate in its detections. We choose YOLOv2 for the purpose of detecting different colored cones. Thresholds for it are chosen such that false positives, incorrect detections and misclassification are avoided at any cost; even if that translate to not being able to detect all cones in a given image. We customized YOLOv2 by reducing the number of classes that it detects, as ``gotthard driverless'' does not really care about detecting cats, dogs, airplanes or bikes to name a few but needs to distinguish and detect `yellow', `blue' and `orange' cones that provide information about the track. We reduce the classes of the pre-trained YOLOv2 on PASCAL-VOC from 20 to 3.

Data for training and testing was initially acquired via smart-phone cameras, during the required lead time before procuring the Basler ``ace'' cameras. The images were manually labeled using a self-developed labeling tool that exploits similar structure between consecutive frames in a video sequence. Objects of interest are easily labeled by drawing rectangle through click-drag-unclick using a mouse. To speed up the tedious labeling procedure, the tool tracks annotated rectangles over frames and propagates them to prevent re-labeling for future frames, treating propagated bounding boxes as annotations. After some frames, the trackers may lose their objects due to fast movement or change in view points. At such points, one can refresh and re-label again. Since the annotations for cones are long and thin rectangular bounding boxes, we exploit such prior information by re-calculating the anchor boxes used by YOLOv2. This is done by performing k-means clustering on the aspect-ratio of the rectangle annotations in the dataset and improves the object detector's performance.

\subsection{Training to detect cones}
We train the object detector on 90\% of our `Cones Dataset 2018', about 2700 manually-annotated images with multiple cones and performance is evaluated on 10\% of the data (about 300 unseen images). The training set is augmented on the fly. Since, the hue is important to recognize color correctly, augmentation is performed in terms of brightness and saturation. Also, as described in YOLOv2, multi-scale training is performed, re-sizing the image after every 10 epochs. The detector was fine tuned over the course of the season as more data was acquired and labeled. Amazon's Web Services (AWS) was used for training.

\begin{table}[h!]
\centering
\begin{tabular}{||c| c c c||} 
 \hline
  & Precision & Recall & mean-Average Precision \\ [0.5ex] 
 \hline\hline
 Training & 0.85 & 0.72 & 0.78 \\ 
 \hline
 Testing & 0.84 & 0.71 & 0.76 \\
 \hline
\end{tabular}
\caption{Performance of YOLO for object detection on the `Cones Dataset 2018'.}
\label{training_testing_YOLO}
\end{table}

% focus on speed, memory/GPU vs accuracy
% allowed to miss cones as long as no false positives
% important to distinguish color as compared to last year where there was trackwalk
% color is important for boundary estimation and is the most reliable piece of info from the camera sensor that keeps the car from going out of track limits
% need to be robust in weather conditions and changing lighting
% training and augmentation scheme
% customized network with anchor boxes chosen only from the cones dataset hand labeled
% discuss labeling tool that uses tracking and cite repo
% trained on AWS/cloud
% data set captured from devices when cameras were not there
% train and test split
% fine tuned over the course of the season as more data was acquired and labeled

\section{Keypoint Regression}
As explained in the previous section, with the help of an object detector, one can find multiple objects of interest in a single image. The question here is to go from objects on the image to their positions in 3D. This in itself is not solvable from a single view of the scene, because of ambiguities due to scale. However, since there is prior information about the 3D shape, size and geometry of the cone, one has hope to recover 3D pose from a single measurement.

\begin{figure}[h]
    \centering
    \includegraphics[width=1.0\textwidth]{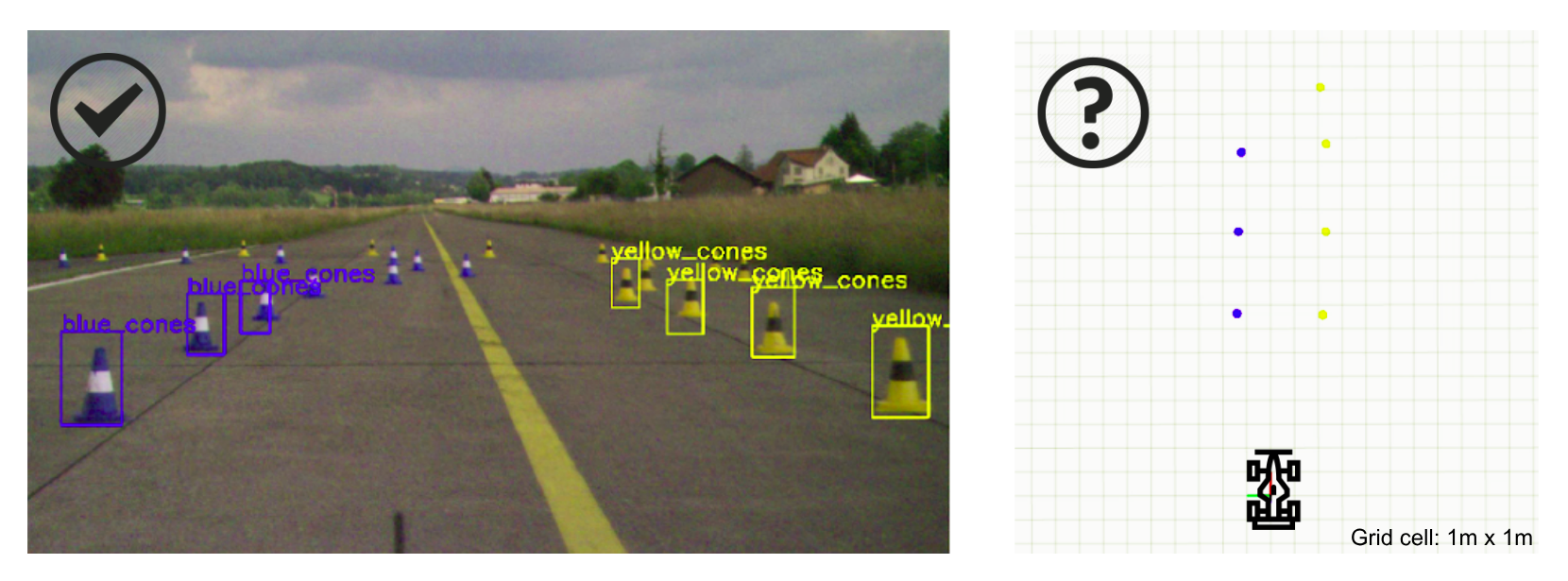}
    \caption{Using an object detector, cones can be detected in an image. However, one needs more information to go from detections on the image to 3D positions. We exploit a priori knowledge about the cone and a calibrated camera to help estimate its depth via 2D-3D correspondences.}
    \label{fig:estimating_pose}
\end{figure}

\subsection{From patches to features - The need for ``keypoint regression''}
Feature extraction has been a key component of computer vision pipeline since a long time. Whether it is panorama stitching, stereo triangulation, structure from motion, place recognition, locating ``points of interest'' that are unique and distinguishable has always been a fundamental technique for decades. There has been a large body of work that strives to find better, faster, more efficient, robust feature extraction techniques. Most of these are very generic and can be used in arbitrary applications. A desirable property that many of these posses is invariance to transformations such as scale, rotation and illumination. Such work includes Harris corners \cite{harris1988combined}, renowned SIFT\cite{lowe2004distinctive}, SURF\cite{bay2006surf}, efficient features with binary descriptors: BRISK\cite{leutenegger2011brisk} and  BRIEF\cite{calonder2010brief}.

The monocular camera's intrinsics are known as a result of the calibration using a large checkerboard to a distance up to 15 meters. One would be able to estimate an object's 3D pose, if there is a 2D-3D correspondence between the 3D object and the 2D image, additionally the calibration parameters of the camera.

The issue when using pre-existing feature extraction techniques is that they are very generic and detect all and any kind of features that follow its criteria. For instance, a Harris corner does not distinguish whether it lies on a cone or on a crack on the asphalt. This makes it hard to draw the relevant correspondences and match them to their 3D counterparts. Another issue is when a patch has a low resolution, it may detect only a couple of features which will not provide enough information to estimate the 3D pose.

To this end, we introduce a feature extraction scheme that is inspired by classical computer vision but has a flavor of learning from data via machine learning.

\subsection{Design and architecture of the ``keypoint regressor''}

\begin{figure}[h]
    \centering
    \includegraphics[width=0.6\textwidth]{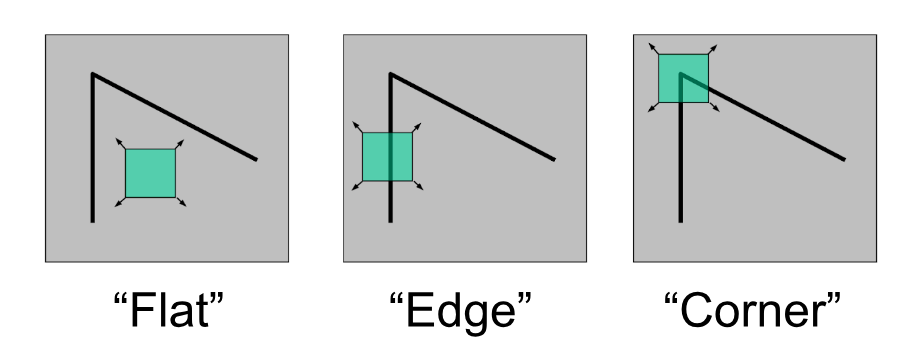}
    \caption{``Flat'' features, ``edges'' and ``corners''. Taken from Seitz, Frolova, Simakov.}
    \label{fig:types_of_features}
\end{figure}

In the context of classical computer vision, there are mainly three kinds of features. The least informative ones are the ``flat features'' which in the vicinity are not distinguishable at all, for instance the patch on a plain, flat wall is one such example. ``Edges'' are a little more interesting as they have a gradient in a particular direction (crossing-over the edge). However, if one moves in a direction perpendicular to this gradient one is unable to distinguish; this is also known as the aperture problem. By far, the most interesting features are the ``corners''. They have change in gradient in two major directions and are quite distinguishable from areas in the vicinity, making them unique and fascinating.

With this in mind, we design a convolutional neural network (CNN) inspired by finding ``corner'' like points given a patch of the image. The primary difference between this scheme and any other feature extraction process is that this is very specific as compared to commonly used techniques. This does not mean that it cannot be used for other objects. This ``keypoint regression'' scheme works for a specific object but can be easily extended to different types of objects. In our case, we want to find position of very specific points on the image that correspond to 3D counterparts whose locations can be measured in 3D from an arbitrary world frame $\mathcal{F}_w$.

\begin{figure}[h]
    \centering
    \includegraphics[width=0.6\textwidth]{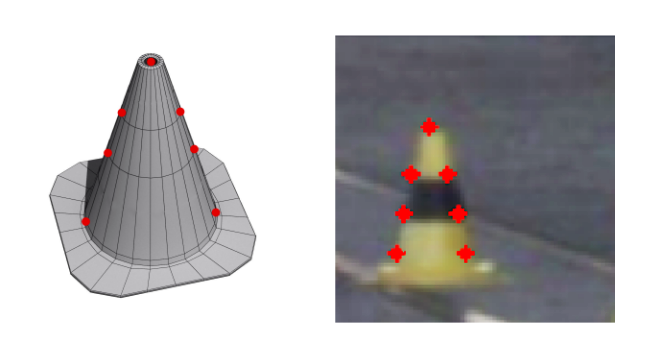}
    \caption{3D model of the cone and a representative sub-image patch with the image of the cone. The red markers correspond to the 7 specific ``keypoints'' the ``keypoint network'' regresses to given an image patch with a cone in it.}
    \label{fig:model_and_kp}
\end{figure}

As depicted in Figure \ref{fig:model_and_kp}, the keypoints on the 3D and its corresponding 2D image are very specific. There are two primary reasons to have these keypoints at those places. First, the keypoint regressor locates position of 7 very specific features that are also visually distinct and can be considered as ``corners'' such as points between the merging of distinct textures and points at the interface of the foreground and the background. Second, and more importantly, these 7 points are relatively easy to measure in 3D from a fixed world frame $\mathcal{F}_w$. For convenience $\mathcal{F}_w$ is chosen to be the base of the 3D cone, enabling easy measurement of 3D position of these 7 points in this world frame, $\mathcal{F}_w$. The 7 keypoints are the apex of the cone, two points (one on either side) at the base of the cone, 4 points where the center stripe, background and upper or lower stripes meet.

The customized CNN inspired from ``corner'' features takes as input a $80 \times 80 \times 3$ sub-image patch which presumably contains a cone, as detected by the object detector in the previous sub-module and maps it to $\mathbb{R}^{14}$. The input dimensions are chosen as $80 \times 80$ spatially, as this was the average size of bounding boxes detected. The output vector of $\mathbb{R}^{14}$ are the $(x, y)$ coordinates of the 7 keypoints relative to the patch.

The architecture of the convolutional neural network consists of basic residual blocks inspired from ResNet \cite{he2016deep}. The reasoning here is that since the convolution operation reduces spatial dimensions, we apply \textit{`same'} convolutions that result from a $3 \times 3$ kernel with \textit{padding=1} and \textit{stride=1} via a residual block (see Equation \ref{eq:conv_formula} and PyTorch \cite{pytorch} docs for more details). As analyzed in \cite{unet}, with more layers, the tensor volume has more channels and fewer spatial dimensions, implying the tensors contain more generic, global information than specific, local information. Since, we eventually care about location of keypoints which is extremely specific and local. Using such an architecture prevents loss of spatial information as it is crucial to predict the position of keypoints accurately as the input volume is processed deeper into the network. Also, the residual blocks can easily learn identity transforms drastically reducing the chance of over-fitting.

\begin{align}
\begin{aligned}
H_{out} = \left\lfloor\frac{H_{in}  + 2 \times \text{padding}[0] - \text{dilation}[0] \times (\text{kernel\_size}[0] - 1) - 1} {\text{stride}[0]} + 1 \right\rfloor\\
W_{out} = \left\lfloor\frac{W_{in}  + 2 \times \text{padding}[1] - \text{dilation}[1] \times (\text{kernel\_size}[1] - 1) - 1}{\text{stride}[1]} + 1\right\rfloor
\end{aligned}
\label{eq:conv_formula}
\end{align}

\begin{figure}[h]
    \centering
    \includegraphics[width=1.0\textwidth]{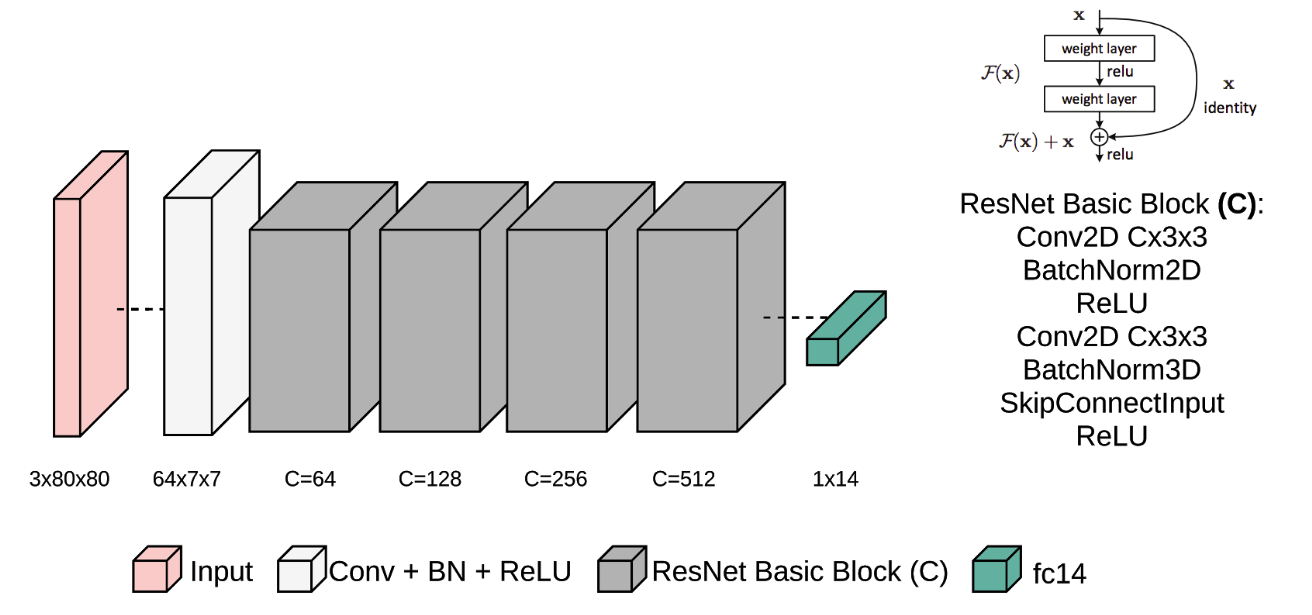}
    \caption{Architecture of the ``keypoint network''. It takes a sub-image patch of $80 \times 80 \times 3$ as input and maps it to $\mathbb{R}^{14}$, the $(x, y)$ coordinates for the 7 keypoints.}
    \label{fig:network_arch}
\end{figure}

The first block in the network is a convolution layer with a batch norm (BN) followed by rectified linear units (ReLU) as the non-linear activation. The next 4 blocks are basic residual blocks with increasing channels $C = 64$, $C = 128$, $C = 256$ and $C = 512$ as depicted in Figure\ref{fig:network_arch}. Finally, there is a fully-connected layer that regresses the location of the keypoints.

\subsection{Loss function}
As briefly mentioned earlier and will be glossed over in detail in a later section, we use object-specific prior information to match 2D-3D correspondences, from the keypoints to a physical cone. The ``keypoint network'' also exploits a priori information about the object's 3D geometry and appearance through the loss function. It uses the concept of the \textit{cross-ratio}.

\subsubsection{The cross ratio}
An intriguing aspect of transforms such as a similarity or an affinity are the invariance under such transforms. A projective transform is more complex than a similarity or an affinity and neither distances between points nor the ratio of these distance is preserved under a projective transform. However, a more complicated entity known as the \textit{cross-ratio}, which is the ratio of ratio of distances, is invariant and remains constant under a projection. Although, a very sophisticated property, the cross ratio has seldom been used in recent works.

The cross-ratio ($Cr$) is a scalar quantity and can be calculated using 4 collinear points or 5 or more non-collinear points \cite{crossratio}. Since it is invariant under a projection and a camera in essence is a projective transform, this implies that the cross-ratio is preserved. It is preserved irrespective of the view point of the scene and whether it is calculated in 3D or in 2D (on the image plane, after the projective transform).

In our case, we use 4 collinear points $p_1, p_2, p_3, p_4$ to calculate the cross-ratio as defined in Equation \ref{eq:cross_ratio}. Depending on whether the value is calculated for 3D points ($D=3$) or their projected 2D counterparts ($D=2$), the distance $\Delta_{ij}$, between two points $p_i$ and $p_j$ is defined.

% $ \Sigma_{i=1}^{7} (p_i^{(x)} - p_{i\_groundtruth}^{(x)})^2 + (p_i^{(y)} - p_{i\_groundtruth}^{(y)})^2 $

% $ + (Cr(p_1, p_2, p_3, p_4) - Cr_{3D})^2 $

% $ + (Cr(p_1, p_5, p_6, p_7) - Cr_{3D})^2 $

% $ Cr(p_1, p_2, p_3, p_4) = (\Delta_{13}/\Delta_{14})/(\Delta_{23}/\Delta_{24}) \in \mathbb{R}$

% $ \Delta_{ij} = \sqrt{\Sigma_{n=1}^{D} (x_{i}^{(n)} - x_{j}^{(n)})^2}, D \in \{2, 3\}$ 

\begin{align}
\begin{aligned}
Cr(p_1, p_2, p_3, p_4) = (\Delta_{13}/\Delta_{14})/(\Delta_{23}/\Delta_{24}) \in \mathbb{R}\\
\Delta_{ij} = \sqrt{\Sigma_{n=1}^{D} (x_{i}^{(n)} - x_{j}^{(n)})^2}, D \in \{2, 3\}
\end{aligned}
\label{eq:cross_ratio}
\end{align}

\subsubsection{Jointly minimizing the squared error and the cross ratio}

In addition to the cross-ratio to act as a regularizer, the loss has a squared error term. This forces the output to be as close as possible to the ground-truth annotation of the keypoints. The effect of the cross-ratio is controlled by the factor $\gamma$.

\begin{align}
\begin{aligned}
\Sigma_{i=1}^{7} (p_i^{(x)} - p_{i\_groundtruth}^{(x)})^2 + (p_i^{(y)} - p_{i\_groundtruth}^{(y)})^2 \\
+ \gamma (Cr(p_1, p_2, p_3, p_4) - Cr_{3D})^2 \\
+ \gamma (Cr(p_1, p_5, p_6, p_7) - Cr_{3D})^2
\end{aligned}
\label{eq:loss_fn}
\end{align}

\begin{figure}[h]
    \centering
    \includegraphics[width=0.4\textwidth]{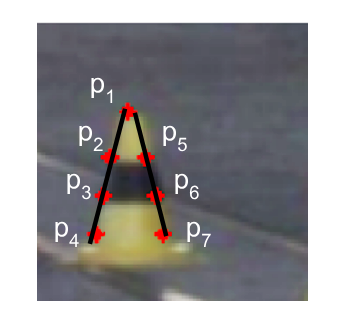}
    \caption{An exemplary $80 \times 80$ cone patch with regressed ``keypoints'' overlaid in red. Depiction of the left ($p_1, p_2, p_3, p_4$) and right arm ($p_1, p_5, p_6, p_7$) of the cone. Both of which are used to calculate the cross-ratio terms and minimize the error between themselves and the cross-ratio on the 3D object ($Cr_{3D}$).}
    \label{fig:cross_ratio_on_cone}
\end{figure}

The second and third term minimize the error between the cross-ratio measured in 3D ($Cr_{3D}$) and the cross-ratio calculated in 2D based on the ``keypoint regressor's'' output, indirectly having an influence on the locations output by the CNN. The second term in Equation \ref{eq:loss_fn} represents the left arm of the cone while the third term is for the right arm, as illustrated in Figure \ref{fig:cross_ratio_on_cone}. For the cross-ratio, we choose to minimize the squared error term between the already known 3D estimate ($Cr_{3D}=1.3940842428872968$) and its 2D counterpart.

Equation \ref{eq:loss_fn} represents the loss function minimized while training the ``keypoint regressor''. The training scheme is explained in the following section.

\subsection{Training scheme}
Cone patches were extracted from full images and manually hand-labeled. The dataset was further augmented by transforming the image with 20 random transforms consisting of rotation, scaling and translation. The data is split as 16,000 cone patches for training and 2,000 cone patches for testing. During the training procedure, the data is further augmented on the fly in the form of contrast, saturation and brightness. Stochastic Gradient Descent (SGD) was used for optimization, with a learning rate, $lr=0.0001$ and $momentum=0.9$ and a batch size of 128. The learning rate is scaled by 0.1 after the first 75 and 100 epochs. The network is trained for 250 epochs. The ``keypoint regressor'' is implemented in PyTorch and used via ROS's \texttt{rospy} interface on ``gotthard driverless.

\begin{figure}[h]
    \centering
    \includegraphics[width=0.4\textwidth]{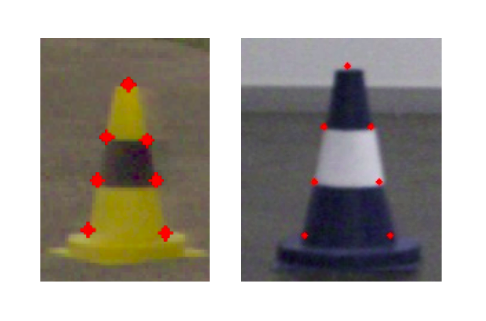}
    \caption{Keypoints regressed on cone patches using the ``keypoint regressor''.}
    \label{fig:ex_keypoints}
\end{figure}

\begin{table}[h!]
\centering
\begin{tabular}{||c| c||} 
 \hline
  & Loss \\ [0.5ex] 
 \hline\hline
 Training & 3.535 $px^2$ \\ 
 \hline
 Testing & 3.783 $px^2$  \\
 \hline
\end{tabular}
\caption{Performance of ``keypoint regressor'' on training and testing datasets.}
\label{training_testing_kp}
\end{table}

\section{2D-3D Correspondences and 3D Pose Estimation}
The ``keypoint network'' provides with accurate locations of very specific features, the keypoints. Since, there is a priori information available about the shape, size, appearance and 3D geometry of the object, the cone in this case, 2D-3D correspondences can be matched. With access to a calibrated camera and 2D-3D correspondences, it is possible to estimate the pose of the object in question from a single image.

We define the camera frame as $\mathcal{F}_c$ and the world frame as $\mathcal{F}_w$. $\mathcal{F}_w$ can be chose arbitrarily, as long as it is used consistently. In this case, we choose the world frame, $\mathcal{F}_w$ to be at the base of the cone, for ease of measurement of the 3D location of the keypoints (with respect to $\mathcal{F}_w$) and convenience of calculation, as will become apparent.

\begin{figure}[h]
    \centering
    \includegraphics[width=0.8\textwidth]{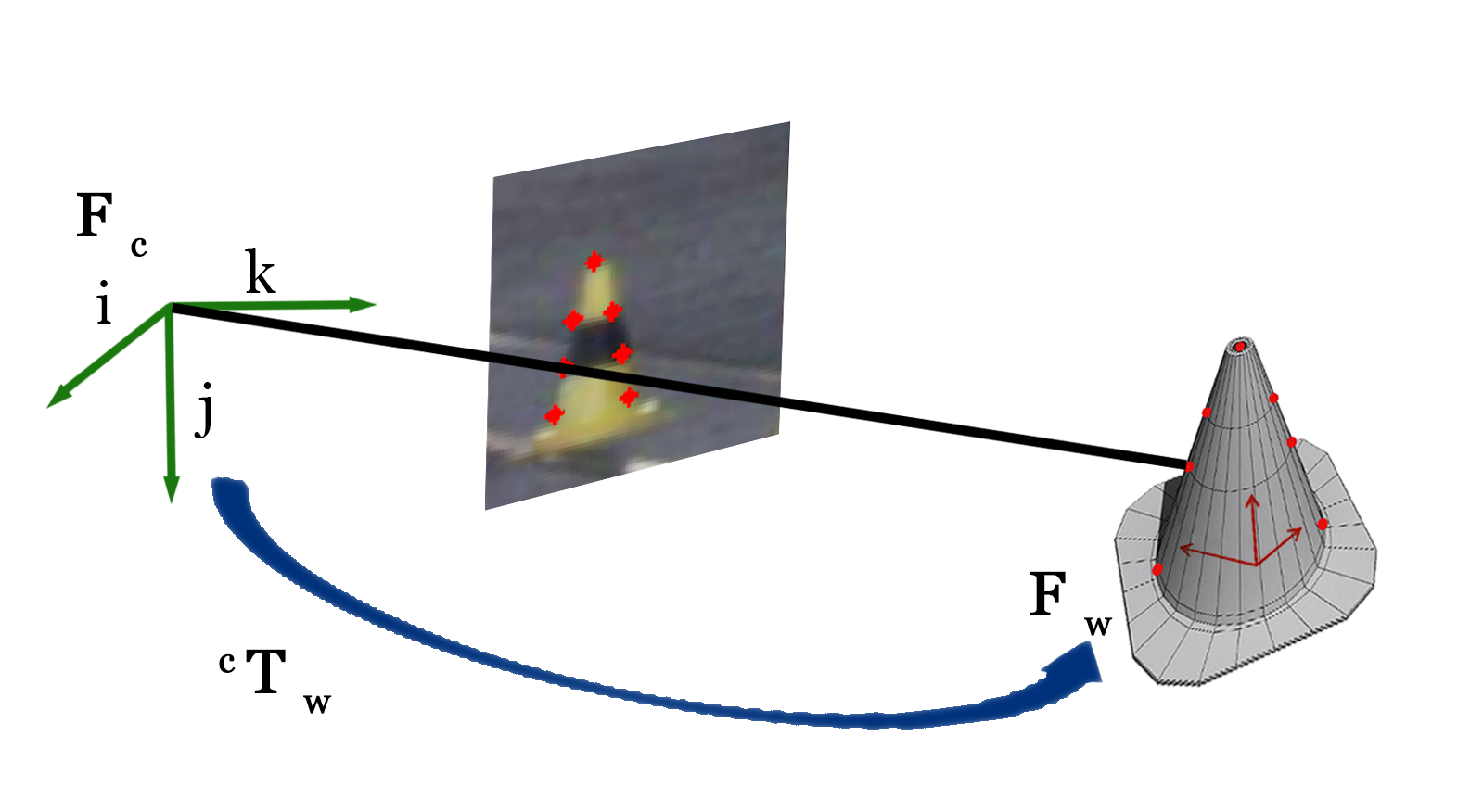}
    \caption{Schematic illustrating matching of 2D-3D correspondence and estimation of transformation between the camera frame and the world frame.}
    \label{fig:cone_pnp}
\end{figure}

We use Perspective n-Point or PnP to estimate the pose of every detected cone. This works by estimating the transform $^{c}\mathcal{T}_w$ between the camera coordinate system and the world coordinate system. Since, the world coordinate system is located at the base of the cone, lying at an arbitrary location (that we want to estimate) in $\mathbb{R}^3$ this transform is exactly the pose we are looking for. A pose consists of a translation and a rotation. The fact that the cone is symmetric along the axis through it's apex and center of the base simplifies the situation. As we are concerned only with the translation between $\mathcal{F}_c$ and $\mathcal{F}_w$, which is exactly the position that we care to estimate, we can discard the orientation due to the cone's symmetric geometry.

To estimate the position of the cone accurately, we use the non-linear version of the PnP implemented in the OpenCV library \cite{opencv3Drecon} that uses Levenberg-Marquardt to obtain the transformation. In addition, RANSAC PnP is used instead of vanilla PnP, tackling noise correspondences. RANSAC PnP can be done for every cone detected, that is extract the 7 features by passing the patch through the ``keypoint regressor'' and use the pre-computed 3D correspondences to estimate the transform, allowing to estimate pose of multiple objects from a single image using a priori knowledge about the object of interest. In the following chapter we discuss the results and analysis of the pipeline's accuracy and performance.

%% ----------------------------------------------------------------------------
% BIWI SA/MA thesis template
%
% Created 09/29/2006 by Andreas Ess
% Extended 13/02/2009 by Jan Lesniak - jlesniak@vision.ee.ethz.ch
%% ----------------------------------------------------------------------------
\newpage
\chapter{Experiments and Results}

\begin{figure}[h]
    \centering
    \includegraphics[width=1.0\textwidth]{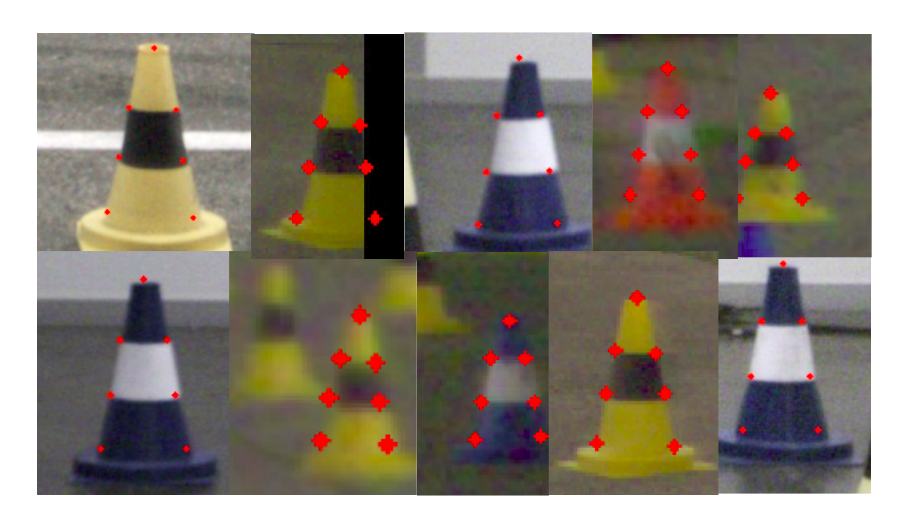}
    \caption{Robust performance of ``keypoint regression''.}
    \label{fig:robust_kp}
\end{figure}

This chapter discusses results and the analysis of the monocular perception pipeline, paying special attention to the ``keypoint network'' and the 3D position estimates of the monocular perception pipeline.

\section{Robust ``keypoint regression''}

In the previous section, the robustness of the first sub-module in the pipeline, the object detector was discussed. It provides good detections in diverse lighting and weather conditions and feeds in a good input to the next sub-module in line, the ``keypoint network''.

Figure \ref{fig:robust_kp} illustrates 10 patches regressed for keypoints after being detected by YOLOv2. There are a couple of interesting cases here. The second cone (from the left) in the top row is one of them. This cone is detected on the right edge of the image and is only partially visible on the image and has almost a third of it truncated (padded with black pixels). Even with missing pixels and no information about a part of the cone, the ``keypoint regressor'' is able to regress correctly. It seems to have learned the geometry and relative location of one keypoint with respect to another. So even when it doesn't observe the cone completely, it is still able to `hallucinate' where the keypoint would have been in case of a complete image. This type of behavior, where it learns the spatial arrangement of the keypoints and their geometry through data and by minimizing the loss is quite intriguing.

Another interesting observation is the second cone (from the left) in the bottom row. Here, there is a blurry cone in the back ground but the keypoint network is able to regress correctly to the most prominent cone, the one in the foreground.

Figure \ref{fig:robust_kp} shows the robustness and accuracy of the ``keypoint regressor'', but that represents only the internal performance of a sub-module. In the following sections, we analyze how outputs of intermediate sub-modules affect the 3D cone positions and how variation in outputs ripples through the pipeline and influence the final output, the position estimates of multiple objects.

\section{3D position accuracy}

\begin{figure}[h]
    \centering
    \includegraphics[width=1.0\textwidth]{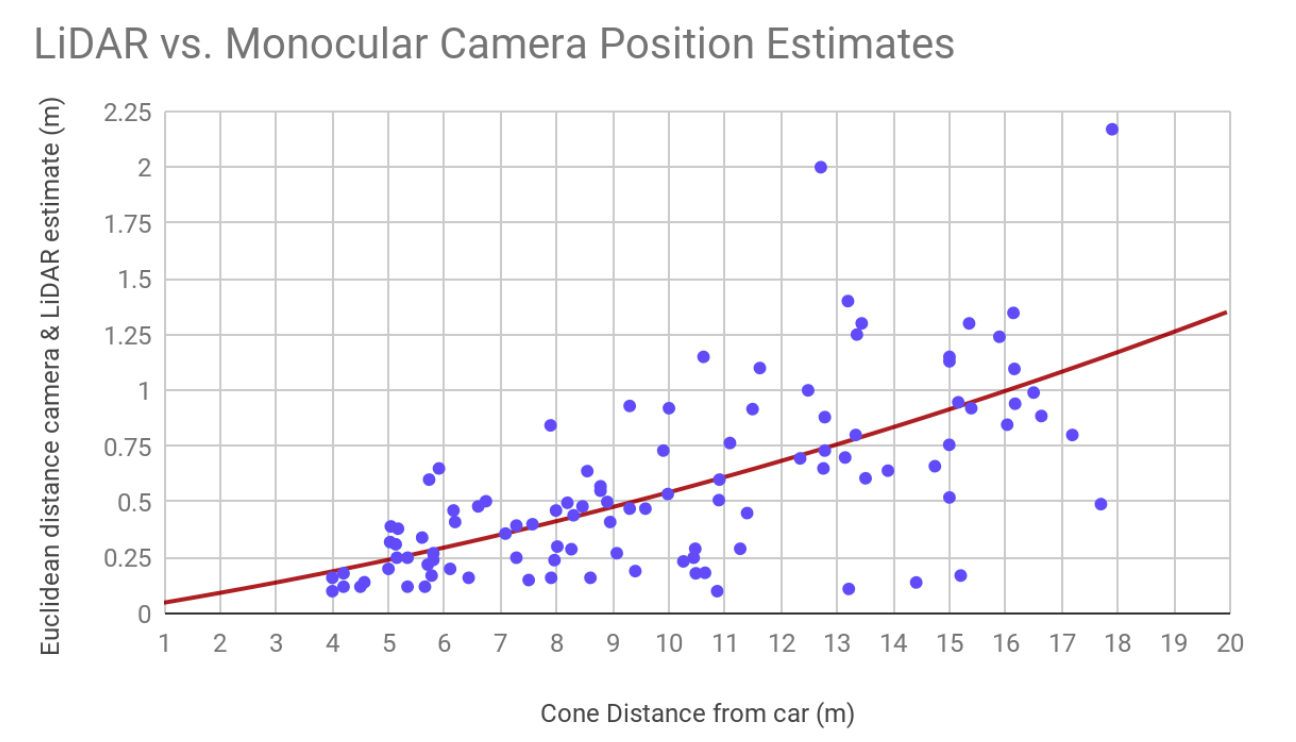}
    \caption{Euclidean distance between position estimates from LiDAR and monocular camera pipeline for the same physical cone. The x-axis represents the distance of the cone from the car and the y-axis the difference between the LiDAR and camera estimates.}
    \label{fig:lidar_mono}
\end{figure}

Apart from color information, one of the most crucial aspect of the pipeline is the accuracy of the 3D pose. Since, this whole approach tries to solve an ill-posed problem by using additional a priori information via shape, size and geometry of the 3D object, it is necessary to compare the position accuracy of the pipeline. In this case, we compare the accuracy with the LiDAR sensor providing ground-truth estimates. Here we assume that the cones detected by LiDAR and their position estimates are accurate to a sub-centimeter scale and treat it as if it were the true position of the cone.

Figure \ref{fig:lidar_mono} is plotted using data from 2 \texttt{rosbags}. The x-axis represents the depth, in meters, of a physical cone and along the y-axis is the Euclidean distance between the 3D position from the LiDAR estimates and the 3D position from the monocular camera pipeline. The plot consists of 104 data points (representing 104 physical cones). Further, a second order curve is fitted to the data, which has mostly linear components. On average, the difference is about 0.5 meters at 10 meters distance away from the car and only about 1 meter at 16 meters. At 5 meters the cone position is off by $\pm 5.00\%$ of its distance, and at 16 meters, it is off by only $\pm 6.25\%$ of its distance.

\section{Extended perception range using the monocular pipeline}

One of the goals of the computer vision perception pipeline was to have an extended range of perception. In previous chapters, we explain in detail how this is achieved. In this section, we show two classical cases of extended range that helps ``gotthard driverless'' anticipate better, allowing for better trajectory and path planning, and more optimal steering and throttle commands from the controller.

\begin{figure}[h]
    \centering
    \includegraphics[width=1.0\textwidth]{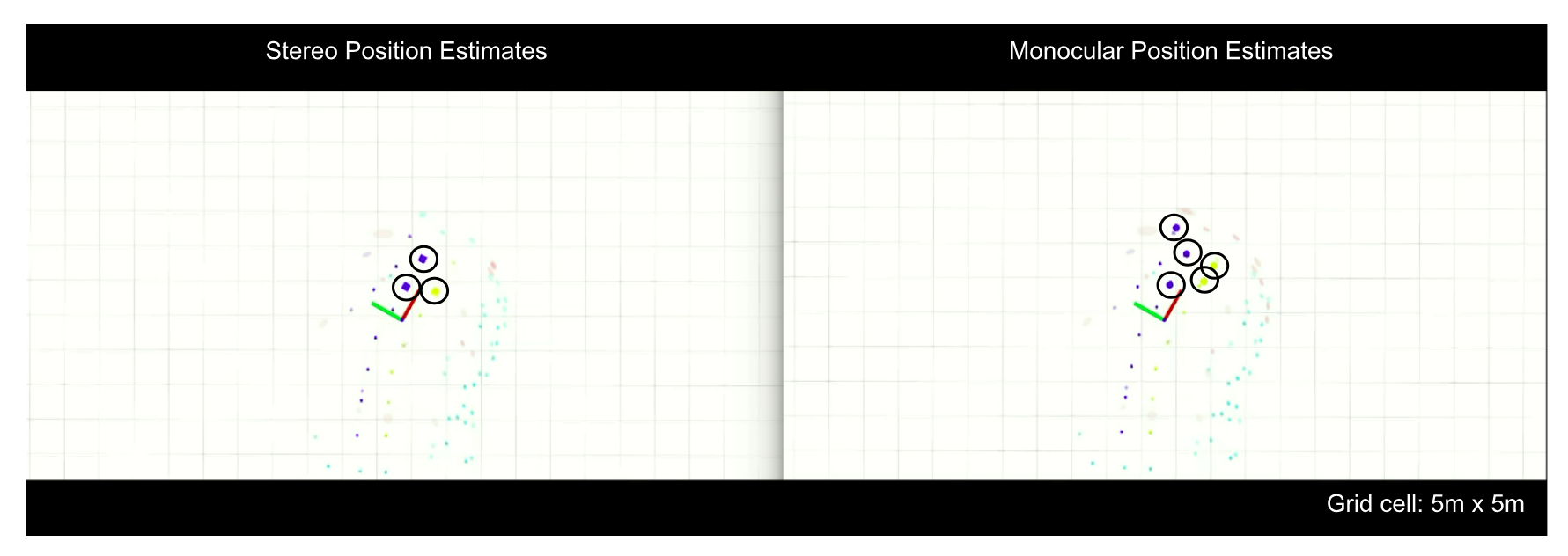}
    \caption{The difference between the range of the stereo and monocular pipeline are clear from this picture. The stereo pipeline accurately triangulates nearby cones, while the monocular pipeline extends the range of perception. At this instance, ``gotthard driverless'' (shown as a coordinate system), approaches a sharp hair-pin turn. It can be observed that the monocular estimates allow to perceive cones on the other side of the track, allowing SLAM and path planning to anticipate better by seeing farther. The grid cell depicted here are 5 meters by 5 meters.}
    \label{fig:mono_range_1}
\end{figure}

Figure \ref{fig:mono_range_1} shows how an extended range of up to 15 meters allows to perceive cones across the track, on the other side. This is very crucial for SLAM and guides the path planning to consider sharp bends or hair-pins. Refer to Figure \ref{fig:mono_range_1} for more details.

\begin{figure}[h!]
    \centering
    \includegraphics[width=1.0\textwidth]{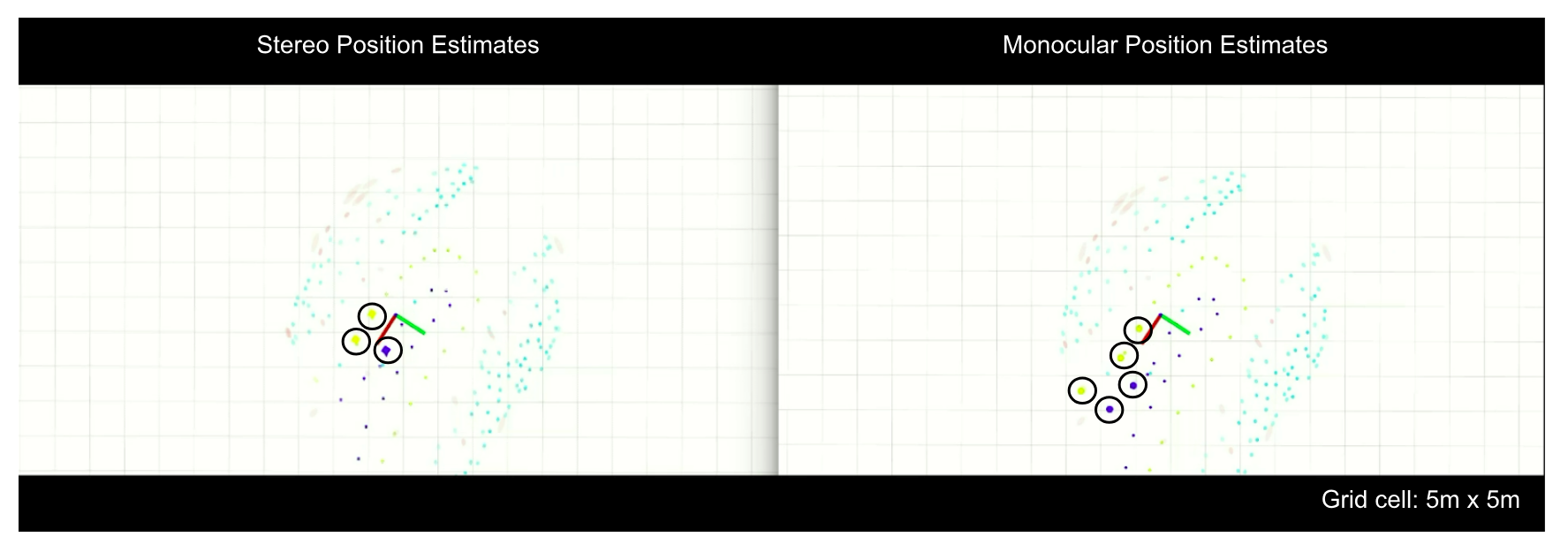}
    \caption{In this case, again, the extended perception range that the monocular pipeline adds on to the perception system on ``gotthard driverless'' is evident. The monocular camera can see and estimate cone positions confidently up to 15 meters. The grid cell depicted here are 5 meters by 5 meters.}
    \label{fig:mono_range_2}
\end{figure}

Another important part is being able to see down a long straight, helping ``gotthard driverless'' to push harder and drive faster. Refer to Figure \ref{fig:mono_range_2} for details.

In Figures \ref{fig:mono_range_1} and \ref{fig:mono_range_2}, we compare the difference between the ranges of the monocular and the stereo pipeline. While the stereo pipeline provides extremely accurate cone positions via triangulation, the monocular pipeline extends the range for the computer vision perception system.

\section{Effect of varying bounding boxes on 3D estimates}

\begin{figure}[h]
    \centering
    \includegraphics[width=0.8\textwidth]{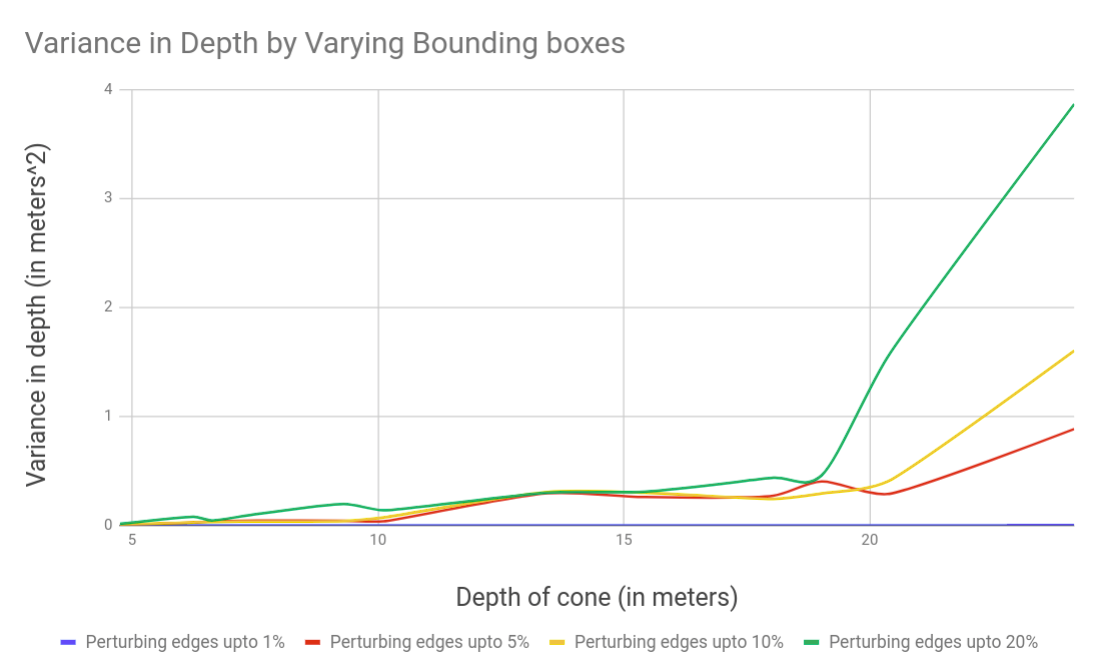}
    \caption{This plot depicts the variance observed in depth of cones when perturbing the bounding boxes that are input to the ``keypoint regressor''. The cone patches are perturbed by various amounts illustrated by different colored lines. On the x-axis is the depth of the cone while the y-axis represents the variance in the cone's depth estimate. This demonstrates that even with noisy and not so precise patches, the ``keypoint regressor'' is able to keep the position variance quite low.}
    \label{fig:obj_det_3D_pose}
\end{figure}

As mentioned before, we would like to see how sub-modules have an effect on the final 3D position estimates. In this experiment, we randomly perturb the edges by an amount proportional to the height and width of the bounding box in respective directions. Perturbation of edges corresponds to randomly moving them up or down and left or right. Estimating depth is most challenging using raw data from cameras. Figure \ref{fig:obj_det_3D_pose} shows how for single images, perturbing the boxes by a certain amount ($\pm 1\%$, $\pm 5\%$, $\pm 10\%$ and $\pm 20\%$) influences the variance in depth estimates. As expected, for higher amounts of perturbation, more variance in depth estimates is observed. However, even for a $\pm 20\%$ perturbation, the variance is about 1 meter$^2$ at 15 meters which is the maximum distance the cameras intrinsics are calibrated for. This perturbation directly affects the pipeline as it is the input to the ``keypoint regression''. This plot shows that even with inaccurate and changing bounding boxes, the 3D cone position, especially the depth is consistent and has low variance.

\section{Effect of varying keypoints on 3D estimates}
Figure \ref{fig:obj_det_3D_pose} shows the affect of imprecise bounding boxes on the 3D depth estimates which are extremely crucial for a driverless race-car. In this section, we discuss how variance in keypoint positions (both x and y coordinates on the image) translates to variance in 3D depth estimates.

\begin{figure}[h]
    \centering
    \includegraphics[width=0.8\textwidth]{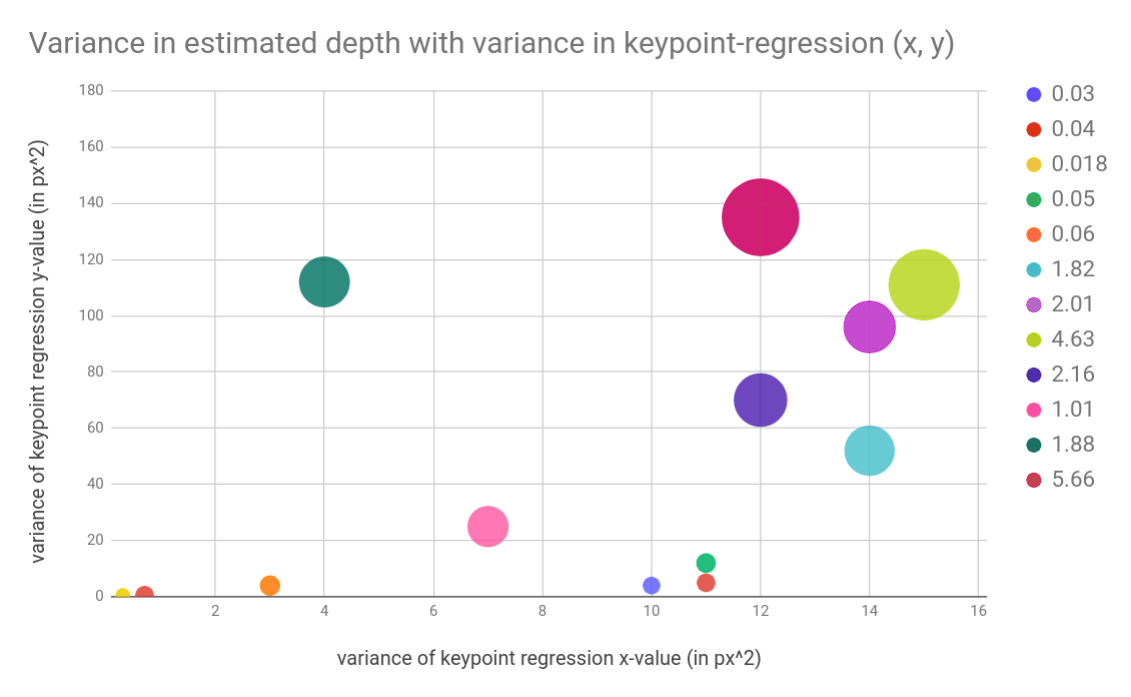}
    \caption{The effect of variance in keypoints (x and y coordinates, on x-axis and y-axis respectively) on the variance in 3D depth of a cone. The variance in depth is depicted by the size of the circles in this bubble chart.}
    \label{fig:kp_var_3D_pose}
\end{figure}

The x-axis in Figure \ref{fig:kp_var_3D_pose} depicts the variance in regressed keypoints' x-coordinate while the y-axis represents the variance in regressed keypoints' y-coordinate. This plot contains data for 12 different physical cones, each represented as a circle on the plot. The size of the circle is proportional to its variance in 3D depth (along the viewing ray). The variance is also mentioned in the legend in the same figure.

An interesting point to note here is that variance of keypoints along x-axis in the image is quite low as compared to the y-axis. Larger variance in the y-coordinates of the keypoints which are along the vertical edge of an image cause a larger variance in depth. This is intuitive because if the keypoints shift vertically in an image, it basically implies the object getting closer or farther. 

% Describe the evaluation you did in a way, such that an independent researcher can repeat it. Cover the following questions:
% \begin{itemize}
%  \item \textit{What is the experimental setup and methodology?} Describe the setting of the experiments and give all the parameters in detail which you have used. Give a detailed account of how the experiment was conducted.
%  \item \textit{What are your results?} In this section, a \emph{clear description} of the results is given. If you produced lots of data, include only representative data here and put all results into the appendix. 
% \end{itemize}

% $ \Sigma_{i=1}^{7} (p_i^{(x)} - p_{i\_groundtruth}^{(x)})^2 + (p_i^{(y)} - p_{i\_groundtruth}^{(y)})^2 $

% $ + (Cr(p_1, p_2, p_3, p_4) - Cr_{3D})^2 $

% $ + (Cr(p_1, p_5, p_6, p_7) - Cr_{3D})^2 $

% $ Cr(p_1, p_2, p_3, p_4) = (\Delta_{13}/\Delta_{14})/(\Delta_{23}/\Delta_{24}) \in \mathbb{R}$

% $ \Delta_{ij} = \sqrt{\Sigma_{n=1}^{D} (x_{i}^{(n)} - x_{j}^{(n)})^2}, D \in \{2, 3\}$ 

%% ----------------------------------------------------------------------------
% BIWI SA/MA thesis template
%
% Created 09/29/2006 by Andreas Ess
% Extended 13/02/2009 by Jan Lesniak - jlesniak@vision.ee.ethz.ch
%% ----------------------------------------------------------------------------
\newpage
\chapter{Discussion}

\section{Perception on ``gotthard driverless''}
The pipeline mentioned in this work is used for perception on ``gotthard driverless''. With a clever choice of optics and the ``keypoint regressor'' as well as 2D-3D correspondence both exploiting a priori information in the form of shape, size and 3D geometry of the cone this allows to perceive cones accurately up to 15 meters.

\begin{figure}[h]
    \centering
    \includegraphics[width=0.8\textwidth]{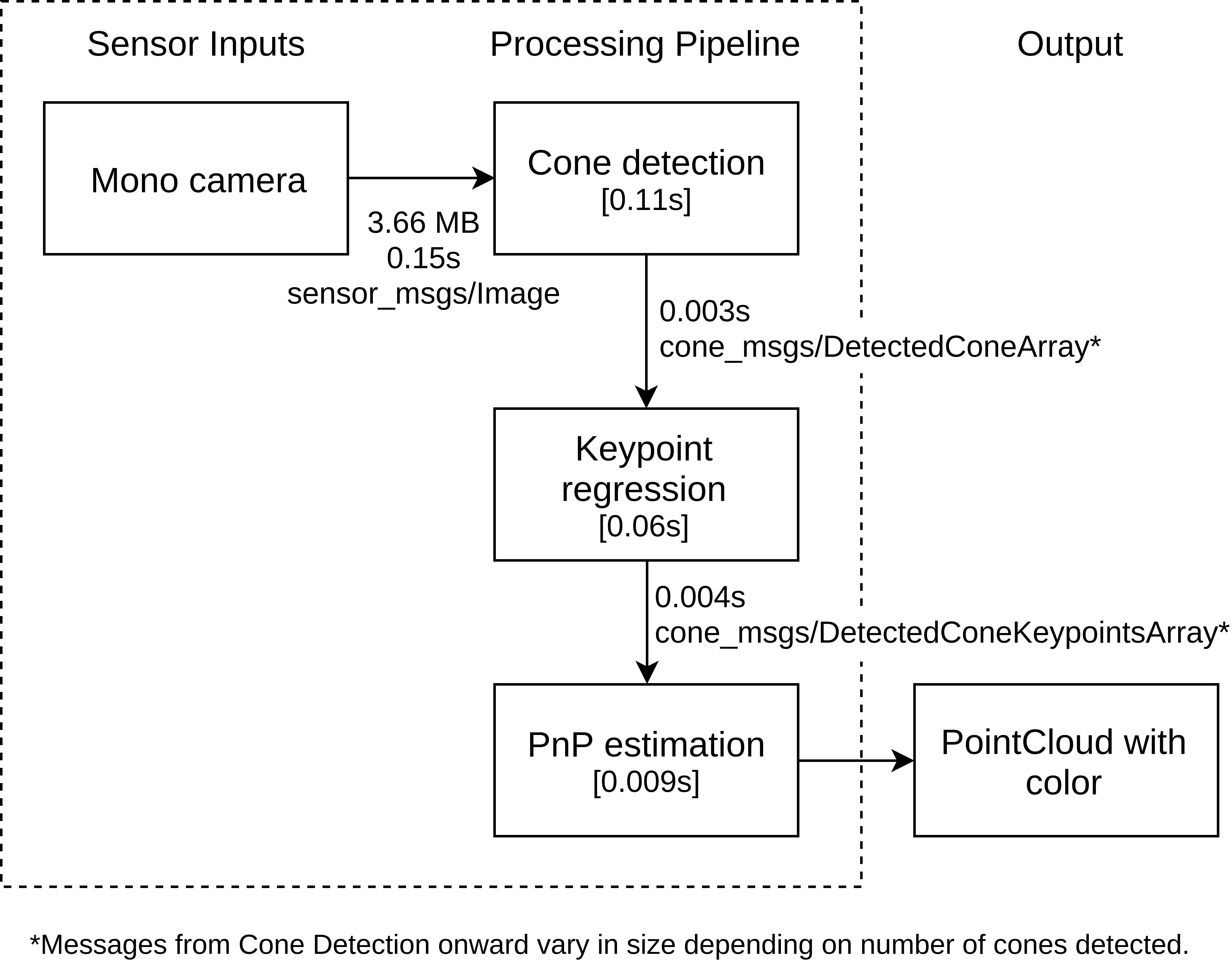}
    \caption{Data flow through the monocular pipeline while running completely on Nvidia's Jetson TX2. It is worth noting that the Jetson TX2 also handles all the raw sensor data from the three cameras directly. This diagram shows the sub-module which represent \texttt{ROS} nodes running with publisher and subscribers to communicate between them. Customized \texttt{ROS} messages are used to transmit data. The input to the pipeline is the raw image with a \texttt{pointcloud} of cones as the pipeline output that is used as input by SLAM.}
    \label{fig:mono_data_flow}
\end{figure}

The pipeline runs at a rate of 8-9Hz on the Nvidia Jetson TX2 device, with the bottle-neck being the object detection pipeline and image transportation from sensor to the computer. For data flow details refer to Figure \ref{fig:mono_data_flow}.

\subsection{Possible improvements to the pipeline}
One of the short-comings was quite a few missed cones from the object detector. Reducing the confidence threshold for YOLOv2 would result in more detections but also increase the likelihood of getting misclassifications or inaccurate detections. An object detector that would detect more than 95\% of the cones would be extremely beneficial.

Another issue that was more related to the positioning of the cameras was the restricted field-of-view of the cameras resulting in them not being able to perceive cones on the inside of corners and sharp turns. One way to remedy this would be have cameras that are not facing forward but are at an angle towards the left and the right. Then the stereo left and right could be used as stand-alone, like the monocular camera and turn them to either side so that cones on tight turns can be seen in the image.

The biggest concern was the delay and processing time in the pipeline. Running at very fast speeds would cause detections to be in the past. Although this would still improve the map with more measurements and color information from the cameras but it would not aid too much in the real-time path planning and seeing cones in-front of the car. There was a total of 0.3s from raw sensor image to 3D cone position output. This effect is mitigated by the fact that the pipeline has a range between 4-15 meters. So even at a speed of 12 m/s, a total time of 0.3s would have no affect on the outputs as the car would have traveled 4m in that time and would still get the first cone in-front of the car and not in the past.

The main source of processing time was transfer of raw image data into the pipeline taking a whooping 0.15s, a half of the total time! The next most time intensive sub-module is the object detector which takes 0.11s. Optimizing these two aspects of the pipeline would reduce the time taken by a huge factor. Running the exact same code on the master computer, the PIP-39 with a GTX 1050i slashed the time to receive the image to 0.11s and the time to process an image by YOLOv2 by a factor of 3 to 0.03s. So running the pipeline with more computation resources would improve its performance and hugely reduce processing time.

\section{Using the ``keypoint regression'' for efficient stereo triangulation}
The pipeline described in this work focuses on 3D position estimation for multiple objects using a single image (from the monocular camera) and exploiting object priors.

But ``gotthard driverless'' is also equipped with a stereo pair. We cleverly use parts of the monocular pipeline to have a more efficient stereo pipeline. A naive approach to stereo triangulation would be extracting features such as corners or more sophisticated features (such as SIFT or BRISK) and performing 2D-2D correspondence matching across the two images captured by the left and the right cameras. To further detect cones, one would have to run 2 instances of YOLOv2 which is computational resource intensive.

Instead, we use the ``keypoint regression'' and PnP on a single image from the left camera to acquire 3D position of detected cones. This 3D position is further improved via additional information in the form of a second image of the same scene (captured at the same time instance) from the right camera. The position accuracy is improved by performing triangulation.

Instead of running 2 instances of YOLOv2, bounding boxes are cleverly propagated from the left image to the right image, using only a single instance of YOLOv2 (running on images from the left camera). The 3D position estimated from the left camera image is transformed to the right camera's coordinate frame using extrinsic calibration and stereo geometry between the two cameras. Finally, the 3D position (now in the right camera frame) is projected to propagate bounding box information from the left frame to the right frame. By virtue of bounding box propagation, finding matching cones in the two images is bypassed, which could have been a sticky situation due to identical appearance of cones on the track.

Once bounding boxes around the same physical cone are known for the left and right frames, SIFT features are then extracted. Instead of naively extracting SIFT features on the complete image, they are done only for a pair of bounding boxes, one each from the left and the right image. Features between the two corresponding cone patches are matched and triangulation is performed, resulting in a drastic reduction in search space to match 2D-2D feature correspondences. Instead of taking a mean of the triangulated points, the median is taken in order to obtain robust 3D cone position.

To summarize, stereo triangulation is made efficient by using the monocular in the following way.

\begin{itemize}
    \itemsep-0.5em
    \item Avoid running two instance of GPGPU intensive YOLOv2.
    \item Bypass the need to search for cone patches that correspond to the same physical cone in both left and right frames.
    \item Limit feature extraction only to sub-images that contain cones; instead of naively finding features in the whole image which would be computationally expensive and wasteful as most of them would remain unused.
    \item By virtue of feature matching from propagated bounding boxes (via 3D positions from PnP on regressed keypoints) the search space for matching 2D-2D feature correspondences is drastically reduced, making it efficient and less prone to noise as well as incorrect feature matches.
\end{itemize}

% The discussion section gives an interpretation of what you have done \cite{day2006wap}:

% \begin{itemize}
%  \item \textit{What do your results mean?} Here you discuss, but you do not recapitulate results. Describe principles, relationships and generalizations shown. Also, mention inconsistencies or exceptions you found.
%  \item \textit{How do your results relate to other's work?} Show how your work agrees or disagrees with other's work. Here you can rely on the information you presented in the ``related work'' section.
%  \item \textit{What are implications and applications of your work?} State how your methods may be applied and what implications might be. 
% \end{itemize}

% \noindent Make sure that introduction/related work and the discussion section act as a pair, i.e. ``be sure the discussion section answers what the introduction section asked'' \cite{day2006wap}. 

%% ----------------------------------------------------------------------------
% BIWI SA/MA thesis template
%
% Created 09/29/2006 by Andreas Ess
% Extended 13/02/2009 by Jan Lesniak - jlesniak@vision.ee.ethz.ch
%% ----------------------------------------------------------------------------

\chapter{Conclusion}

We introduce an effective convolutional neural network architecture for ``keypoint regression'' which is capable of running real-time using low-computation power. Prior information about the 3D object is exploited and injected in two ways in the proposed pipeline. First, in the form of the 3D correspondences that allow to measure 3D position from a single frame. Second, this work also introduces a novel loss function with a \textit{cross-ratio} term (again, exploiting 3D geometry of the object), making the complete package lie at the intersection of classical computer vision and data driven deep learning.

The pipeline with all its sub-modules is quite robust to varying conditions and incomplete data. The object detection which is the first sub-module is able to detect cones in diverse conditions including both lighting and weather. Even with truncated cones, missing information and imprecise bounding boxes, the ``keypoint regression'' reliably extracts very specific features accurately. Finally, a RANSAC PnP scheme is used to robustly estimate the 3D position of the detected cones from a single image.

As discussed in the results and analysis from the previous chapters, the monocular pipeline extends the perception range of the computer vision perception system. The 3D estimates from the monocular pipeline are accurate and comparable with a range sensor such as the LiDAR. The 3D position estimates are extremely close between the two, showing the effectiveness of our approach using a monocular camera and a per-frame multi-object detection and 3D position estimation.

The computer vision perception pipeline is designed from scratch. Starting with a blue print and diverse concepts on how to perceive and accurately estimate 3D cone positions. Finally, freezing to this concept and implementing it to run real-time on an embedded Jetson TX2 computing platform.

Not only do the results hold in theory and on paper but the effectiveness and accuracy of the pipeline is reflected in ``gotthard driverless'''s performance on the track. It participated in 2 driverless events, Formula Student Italy and Formula Student Germany.

At Formula Student Italy held at Varano de Melegari between 11-15 July, 2018, ``gotthard driverless'' won all the static and dynamic events, scoring the highest ever and the first 1000/1000 points in any Formula Student event in the history.

With some more testing time after FS Italy, at Formula Student Germany held at Hockenheimring, Germany from 6-12 August, 2018, ``gotthard driverless'' asserted its FSI performance by being crowned as Champion once again. We were finished first in Engineering Design, Trackdrive, Skid-pad and Efficiency. We scored the highest ever points by a team at FS Germany, 959.57/1000.

To checkout our performance in the track-drive at Formula Student Germany, 2018, with ``gotthard'' cruising at 54 kmph autonomously, click on this link. \url{https://youtu.be/HegmIXASKow?t=11694}

%% ----------------------------------------------------------------------------
% If Appendix is needed
%% ----------------------------------------------------------------------------
% \appendix
% \input{appendix.tex}

%% ----------------------------------------------------------------------------
% Bibliography is stored in references.bib file, and can often be found
% online on webpages like dblp.uni-trier.de
%
% To include it in your thesis, run
%  pdflatex main
%  bibtex main
%  pdflatex main
%  pdflatex main
%
% This ensures all references are done correctly.
%% ----------------------------------------------------------------------------

\bibliographystyle{plain}
\bibliography{references}

\begin{thebibliography}{10}

\bibitem{crossratio}
{Cross ratio}.
\newblock \url{http://robotics.stanford.edu/~birch/projective/node10.html}.
\newblock Accessed: 2018-09-11.

\bibitem{darpaGC}
{DARPA} grand challenge.
\newblock \url{https://en.wikipedia.org/wiki/DARPA\_Grand\_Challenge}.
\newblock Accessed: 2018-09-11.

\bibitem{opencv3Drecon}
{OpenCV} calibration and 3d reconstruction.
\newblock
  \url{https://docs.opencv.org/2.4/modules/calib3d/doc/camera\_calibration\_and\_3d\_reconstruction.html#solvepnp}.
\newblock Accessed: 2018-09-11.

\bibitem{pytorch}
{PyTorch}.
\newblock \url{https://pytorch.org/}.

\bibitem{ros}
{ROS: Robot Operating System}.
\newblock \url{https://ros.org/}.

\bibitem{bay2006surf}
Herbert Bay, Tinne Tuytelaars, and Luc Van~Gool.
\newblock Surf: Speeded up robust features.
\newblock In {\em European conference on computer vision}, pages 404--417.
  Springer, 2006.

\bibitem{nvidiaBojarskiTDFFGJM16}
Mariusz Bojarski, Davide~Del Testa, Daniel Dworakowski, Bernhard Firner, Beat
  Flepp, Prasoon Goyal, Lawrence~D. Jackel, Mathew Monfort, Urs Muller, Jiakai
  Zhang, Xin Zhang, Jake Zhao, and Karol Zieba.
\newblock End to end learning for self-driving cars.
\newblock {\em CoRR}, abs/1604.07316, 2016.

\bibitem{calonder2010brief}
Michael Calonder, Vincent Lepetit, Christoph Strecha, and Pascal Fua.
\newblock Brief: Binary robust independent elementary features.
\newblock In {\em European conference on computer vision}, pages 778--792.
  Springer, 2010.

\bibitem{dalal2005histograms}
Navneet Dalal and Bill Triggs.
\newblock Histograms of oriented gradients for human detection.
\newblock In {\em Computer Vision and Pattern Recognition, 2005. CVPR 2005.
  IEEE Computer Society Conference on}, volume~1, pages 886--893. IEEE, 2005.

\bibitem{fluela}
Miguel de~la Iglesia~Valls, Hubertus Franciscus~Cornelis Hendrikx, Victor
  Reijgwart, Fabio~Vito Meier, Inkyu Sa, Renaud Dub{\'{e}}, Abel~Roman Gawel,
  Mathias B{\"{u}}rki, and Roland Siegwart.
\newblock Design of an autonomous racecar: Perception, state estimation and
  system integration.
\newblock {\em CoRR}, abs/1804.03252, 2018.

\bibitem{imagenet}
Jia Deng, Wei Dong, Richard Socher, Li-Jia Li, Kai Li, and Li~Fei-Fei.
\newblock Imagenet: A large-scale hierarchical image database.
\newblock In {\em Computer Vision and Pattern Recognition, 2009. CVPR 2009.
  IEEE Conference on}, pages 248--255. Ieee, 2009.

\bibitem{felzenszwalb2010object}
Pedro~F Felzenszwalb, Ross~B Girshick, David McAllester, and Deva Ramanan.
\newblock Object detection with discriminatively trained part-based models.
\newblock {\em IEEE transactions on pattern analysis and machine intelligence},
  32(9):1627--1645, 2010.

\bibitem{girshick2015fast}
Ross Girshick.
\newblock Fast r-cnn.
\newblock In {\em Proceedings of the IEEE international conference on computer
  vision}, pages 1440--1448, 2015.

\bibitem{girshick2014rich}
Ross Girshick, Jeff Donahue, Trevor Darrell, and Jitendra Malik.
\newblock Rich feature hierarchies for accurate object detection and semantic
  segmentation.
\newblock In {\em Proceedings of the IEEE conference on computer vision and
  pattern recognition}, pages 580--587, 2014.

\bibitem{gkioxari2014using}
Georgia Gkioxari, Bharath Hariharan, Ross Girshick, and Jitendra Malik.
\newblock Using k-poselets for detecting people and localizing their keypoints.
\newblock In {\em Proceedings of the IEEE Conference on Computer Vision and
  Pattern Recognition}, pages 3582--3589, 2014.

\bibitem{glasner2012aware}
Daniel Glasner, Meirav Galun, Sharon Alpert, Ronen Basri, and Gregory
  Shakhnarovich.
\newblock aware object detection and continuous pose estimation.
\newblock {\em Image and Vision Computing}, 30(12):923--933, 2012.

\bibitem{harris1988combined}
Chris Harris and Mike Stephens.
\newblock A combined corner and edge detector.
\newblock Citeseer, 1988.

\bibitem{he2016deep}
Kaiming He, Xiangyu Zhang, Shaoqing Ren, and Jian Sun.
\newblock Deep residual learning for image recognition.
\newblock In {\em Proceedings of the IEEE conference on computer vision and
  pattern recognition}, pages 770--778, 2016.

\bibitem{lepetit2009epnp}
Vincent Lepetit, Francesc Moreno-Noguer, and Pascal Fua.
\newblock Epnp: An accurate o (n) solution to the pnp problem.
\newblock {\em International journal of computer vision}, 81(2):155, 2009.

\bibitem{leutenegger2011brisk}
Stefan Leutenegger, Margarita Chli, and Roland~Y Siegwart.
\newblock Brisk: Binary robust invariant scalable keypoints.
\newblock In {\em Computer Vision (ICCV), 2011 IEEE International Conference
  on}, pages 2548--2555. IEEE, 2011.

\bibitem{liu2016ssd}
Wei Liu, Dragomir Anguelov, Dumitru Erhan, Christian Szegedy, Scott Reed,
  Cheng-Yang Fu, and Alexander~C Berg.
\newblock Ssd: Single shot multibox detector.
\newblock In {\em European conference on computer vision}, pages 21--37.
  Springer, 2016.

\bibitem{lowe2004distinctive}
David~G Lowe.
\newblock Distinctive image features from scale-invariant keypoints.
\newblock {\em International journal of computer vision}, 60(2):91--110, 2004.

\bibitem{mur2015orb}
Raul Mur-Artal, Jose Maria~Martinez Montiel, and Juan~D Tardos.
\newblock Orb-slam: a versatile and accurate monocular slam system.
\newblock {\em IEEE Transactions on Robotics}, 31(5):1147--1163, 2015.

\bibitem{redmon2016you}
Joseph Redmon, Santosh Divvala, Ross Girshick, and Ali Farhadi.
\newblock You only look once: Unified, real-time object detection.
\newblock In {\em Proceedings of the IEEE conference on computer vision and
  pattern recognition}, pages 779--788, 2016.

\bibitem{YOLOv2}
Joseph Redmon and Ali Farhadi.
\newblock {YOLO9000:} better, faster, stronger.
\newblock {\em CoRR}, abs/1612.08242, 2016.

\bibitem{ren2015faster}
Shaoqing Ren, Kaiming He, Ross Girshick, and Jian Sun.
\newblock Faster r-cnn: Towards real-time object detection with region proposal
  networks.
\newblock In {\em Advances in neural information processing systems}, pages
  91--99, 2015.

\bibitem{unet}
Olaf Ronneberger, Philipp Fischer, and Thomas Brox.
\newblock U-net: Convolutional networks for biomedical image segmentation.
\newblock {\em CoRR}, abs/1505.04597, 2015.

\bibitem{savarese20073d}
Silvio Savarese and Li~Fei-Fei.
\newblock 3d generic object categorization, localization and pose estimation.
\newblock 2007.

\bibitem{vp_and_kp}
Shubham Tulsiani and Jitendra Malik.
\newblock Viewpoints and keypoints.
\newblock {\em CoRR}, abs/1411.6067, 2014.

\bibitem{viola2001rapid}
Paul Viola and Michael Jones.
\newblock Rapid object detection using a boosted cascade of simple features.
\newblock In {\em Computer Vision and Pattern Recognition, 2001. CVPR 2001.
  Proceedings of the 2001 IEEE Computer Society Conference on}, volume~1, pages
  I--I. IEEE, 2001.

\bibitem{posecnn}
Yu~Xiang, Tanner Schmidt, Venkatraman Narayanan, and Dieter Fox.
\newblock Posecnn: {A} convolutional neural network for 6d object pose
  estimation in cluttered scenes.
\newblock {\em CoRR}, abs/1711.00199, 2017.

\end{thebibliography}

\end{document}